\documentclass[10pt,journal,compsoc]{IEEEtran}
\ifCLASSINFOpdf
\else
\fi
\usepackage{graphicx}
\usepackage{amsmath}
\usepackage{amssymb}
\usepackage{booktabs}
\usepackage{url}            %
\usepackage{booktabs}       %
\usepackage{amsfonts}       %
\usepackage{nicefrac}       %
\usepackage{microtype}      %
\usepackage[dvipsnames]{xcolor}        %
\usepackage{times}
\usepackage{epsfig}
\usepackage{graphicx}
\usepackage{booktabs}
\usepackage[separate-uncertainty=true]{siunitx}
\usepackage{multirow}
\usepackage{subcaption}
\usepackage{times}
\usepackage{epsfig}
\usepackage[numbers]{natbib}
\usepackage{rotating}
\usepackage{xspace} %
\usepackage[pagebackref=true,breaklinks=true,colorlinks,bookmarks=false]{hyperref}

\usepackage{scalerel}
\usepackage{tikz}
\usetikzlibrary{svg.path}
\definecolor{orcidlogocol}{HTML}{A6CE39}
\tikzset{
  orcidlogo/.pic={
    \fill[orcidlogocol] svg{M256,128c0,70.7-57.3,128-128,128C57.3,256,0,198.7,0,128C0,57.3,57.3,0,128,0C198.7,0,256,57.3,256,128z};
    \fill[white] svg{M86.3,186.2H70.9V79.1h15.4v48.4V186.2z}
                 svg{M108.9,79.1h41.6c39.6,0,57,28.3,57,53.6c0,27.5-21.5,53.6-56.8,53.6h-41.8V79.1z M124.3,172.4h24.5c34.9,0,42.9-26.5,42.9-39.7c0-21.5-13.7-39.7-43.7-39.7h-23.7V172.4z}
                 svg{M88.7,56.8c0,5.5-4.5,10.1-10.1,10.1c-5.6,0-10.1-4.6-10.1-10.1c0-5.6,4.5-10.1,10.1-10.1C84.2,46.7,88.7,51.3,88.7,56.8z};
  }
}
\newcommand\orcidicon[1]{\href{https://orcid.org/#1}{\mbox{\scalerel*{
\begin{tikzpicture}[yscale=-1,transform shape]
\pic{orcidlogo};
\end{tikzpicture}
}{|}}}}

\begin{document}
%
\title{Stochastic Video Prediction with \\ Structure and Motion}
%
%
%
%

\author{Adil Kaan Akan$^\dagger$ \orcidicon{0000-0003-0022-7224},
        Sadra Safadoust \orcidicon{0000-0003-2018-0451},
        Fatma G\"uney \orcidicon{0000-0002-0358-983X} 
\IEEEcompsocitemizethanks{\IEEEcompsocthanksitem All authors are at the Department of Computer Engineering and KUIS AI Center at Ko\c{c} University in Istanbul, Turkey.\protect\\
E-mail: {\{kakan20, ssafadoust20, fguney\}@ku.edu.tr}
}
\thanks{$^\dagger$ Corresponding author.}}

%
%

\markboth{Akan \MakeLowercase{\textit{et al.}}: Stochastic Video Prediction with Structure and Motion}%
{Akan \MakeLowercase{\textit{et al.}}: Stochastic Video Prediction with Structure and Motion}
\IEEEtitleabstractindextext{%
\begin{abstract}
While stochastic video prediction models enable future prediction under uncertainty, they mostly fail to model the complex dynamics of real-world scenes. For example, they cannot provide reliable predictions for scenes with a moving camera and independently moving foreground objects in driving scenarios. The existing methods fail to fully capture the dynamics of the structured world by only focusing on changes in pixels. In this paper, we assume that there is an underlying process creating observations in a video and propose to factorize it into static and dynamic components. We model the static part based on the scene structure and the ego-motion of the vehicle, and the dynamic part based on the remaining motion of the dynamic objects. By learning separate distributions of changes in foreground and background, we can decompose the scene into static and dynamic parts and separately model the change in each. Our experiments demonstrate that disentangling structure and motion helps stochastic video prediction, leading to better future predictions in complex driving scenarios on two real-world driving datasets, KITTI and Cityscapes.
\end{abstract}

\begin{IEEEkeywords}
Stochastic future prediction, stochastic video prediction, video frame prediction, structure and motion, optical flow
\end{IEEEkeywords}} 

\maketitle

\IEEEdisplaynontitleabstractindextext
\newcommand{\Perp}{\perp\!\!\! \perp}
\newcommand{\bK}{\mathbf{K}}
\newcommand{\bX}{\mathbf{X}}
\newcommand{\bY}{\mathbf{Y}}
\newcommand{\bk}{\mathbf{k}}
\newcommand{\bx}{\mathbf{x}}
\newcommand{\by}{\mathbf{y}}
\newcommand{\bhy}{\hat{\mathbf{y}}}
\newcommand{\bty}{\tilde{\mathbf{y}}}
\newcommand{\bG}{\mathbf{G}}
\newcommand{\bI}{\mathbf{I}}
\newcommand{\bg}{\mathbf{g}}
\newcommand{\bS}{\mathbf{S}}
\newcommand{\bs}{\mathbf{s}}
\newcommand{\bM}{\mathbf{M}}
\newcommand{\bw}{\mathbf{w}}
\newcommand{\eye}{\mathbf{I}}
\newcommand{\bU}{\mathbf{U}}
\newcommand{\bV}{\mathbf{V}}
\newcommand{\bW}{\mathbf{W}}
\newcommand{\bn}{\mathbf{n}}
\newcommand{\bv}{\mathbf{v}}
\newcommand{\bwv}{\mathbf{wv}}
\newcommand{\bq}{\mathbf{q}}
\newcommand{\bR}{\mathbf{R}}
\newcommand{\bi}{\mathbf{i}}
\newcommand{\bj}{\mathbf{j}}
\newcommand{\bp}{\mathbf{p}}
\newcommand{\bt}{\mathbf{t}}
\newcommand{\bJ}{\mathbf{J}}
\newcommand{\bu}{\mathbf{u}}
\newcommand{\bB}{\mathbf{B}}
\newcommand{\bD}{\mathbf{D}}
\newcommand{\bz}{\mathbf{z}}
\newcommand{\bP}{\mathbf{P}}
\newcommand{\bC}{\mathbf{C}}
\newcommand{\bA}{\mathbf{A}}
\newcommand{\bZ}{\mathbf{Z}}
\newcommand{\bff}{\mathbf{f}}
\newcommand{\bF}{\mathbf{F}}
\newcommand{\bo}{\mathbf{o}}
\newcommand{\bO}{\mathbf{O}}
\newcommand{\bc}{\mathbf{c}}
\newcommand{\bm}{\mathbf{m}}
\newcommand{\bT}{\mathbf{T}}
\newcommand{\bQ}{\mathbf{Q}}
\newcommand{\bL}{\mathbf{L}}
\newcommand{\bl}{\mathbf{l}}
\newcommand{\ba}{\mathbf{a}}
\newcommand{\bE}{\mathbf{E}}
\newcommand{\bH}{\mathbf{H}}
\newcommand{\bd}{\mathbf{d}}
\newcommand{\br}{\mathbf{r}}
\newcommand{\be}{\mathbf{e}}
\newcommand{\bb}{\mathbf{b}}
\newcommand{\bh}{\mathbf{h}}
\newcommand{\bhh}{\hat{\mathbf{h}}}
\newcommand{\btheta}{\boldsymbol{\theta}}
\newcommand{\bTheta}{\boldsymbol{\Theta}}
\newcommand{\bpi}{\boldsymbol{\pi}}
\newcommand{\bphi}{\boldsymbol{\phi}}
\newcommand{\bpsi}{\boldsymbol{\psi}}
\newcommand{\bPhi}{\boldsymbol{\Phi}}
\newcommand{\bmu}{\boldsymbol{\mu}}
\newcommand{\bsigma}{\boldsymbol{\sigma}}
\newcommand{\bSigma}{\boldsymbol{\Sigma}}
\newcommand{\bGamma}{\boldsymbol{\Gamma}}
\newcommand{\bbeta}{\boldsymbol{\beta}}
\newcommand{\bomega}{\boldsymbol{\omega}}
\newcommand{\blambda}{\boldsymbol{\lambda}}
\newcommand{\bLambda}{\boldsymbol{\Lambda}}
\newcommand{\bkappa}{\boldsymbol{\kappa}}
\newcommand{\btau}{\boldsymbol{\tau}}
\newcommand{\balpha}{\boldsymbol{\alpha}}
\newcommand{\nR}{\mathbb{R}}
\newcommand{\nN}{\mathbb{N}}
\newcommand{\nL}{\mathbb{L}}
\newcommand{\cN}{\mathcal{N}}
\newcommand{\cM}{\mathcal{M}}
\newcommand{\cR}{\mathcal{R}}
\newcommand{\cB}{\mathcal{B}}
\newcommand{\cL}{\mathcal{L}}
\newcommand{\cH}{\mathcal{H}}
\newcommand{\cS}{\mathcal{S}}
\newcommand{\cT}{\mathcal{T}}
\newcommand{\cO}{\mathcal{O}}
\newcommand{\cC}{\mathcal{C}}
\newcommand{\cP}{\mathcal{P}}
\newcommand{\cE}{\mathcal{E}}
\newcommand{\cI}{\mathcal{I}}
\newcommand{\cF}{\mathcal{F}}
\newcommand{\cK}{\mathcal{K}}
\newcommand{\cY}{\mathcal{Y}}
\newcommand{\cX}{\mathcal{X}}
\def\bgamma{\boldsymbol\gamma}

\newcommand{\specialcell}[2][c]{%
  \begin{tabular}[#1]{@{}c@{}}#2\end{tabular}}

\newcommand{\figref}[1]{\Fig~\ref{#1}}
\newcommand{\secref}[1]{Section~\ref{#1}}
\newcommand{\algref}[1]{Algorithm~\ref{#1}}
\newcommand{\eqnref}[1]{Eq.~\eqref{#1}}
\newcommand{\tabref}[1]{Table~\ref{#1}}

\newcommand{\rulesep}{\unskip\ \vrule\ }

\newcommand{\KLD}[2]{D_{\mathrm{KL}} \Big(#1 \mid\mid #2 \Big)}

\renewcommand{\b}{\ensuremath{\mathbf}}

\def\mc{\mathcal}
\def\mb{\mathbf}

\newcommand{\T}{^{\raisemath{-1pt}{\mathsf{T}}}}

\makeatletter
\DeclareRobustCommand\onedot{\futurelet\@let@token\@onedot}
\def\@onedot{\ifx\@let@token.\else.\null\fi\xspace}
\def\eg{e.g\onedot} \def\Eg{E.g\onedot}
\def\ie{i.e\onedot} \def\Ie{I.e\onedot}
\def\cf{cf\onedot} \def\Cf{Cf\onedot}
\def\etc{etc\onedot} \def\vs{vs\onedot}
\def\wrt{wrt\onedot}
\def\dof{d.o.f\onedot}
\def\etal{et~al\onedot} \def\iid{i.i.d\onedot}
\def\Fig{Fig\onedot} \def\Eqn{Eqn\onedot} \def\Sec{Sec\onedot} \def\Alg{Alg\onedot}
\makeatother

\newcommand{\xdownarrow}[1]{%
  {\left\downarrow\vbox to #1{}\right.\kern-\nulldelimiterspace}
}

\newcommand{\xuparrow}[1]{%
  {\left\uparrow\vbox to #1{}\right.\kern-\nulldelimiterspace}
}

\renewcommand\UrlFont{\color{blue}\rmfamily}

\newcommand*\rot{\rotatebox{90}}
\newcommand{\boldparagraph}[1]{\vspace{0.2cm}\noindent{\bf #1:} }
\newcommand{\boldquestion}[1]{\vspace{0.2cm}\noindent{\bf #1} }

\newcommand{\ka}[1]{ \noindent {\color{blue} {\bf Kaan:} {#1}} } 
\newcommand{\ftm}[1]{ \noindent {\color{magenta} {\bf Fatma:} {#1}} }
\newcommand{\aykut}[1]{ \noindent {\color{green} {\bf Aykut:} {#1}} }
\newcommand{\ee}[1]{ \noindent {\color{cyan} {\bf Erkut:} {#1}} }
\newcommand{\safa}[1]{ \noindent {\color{brown} {\bf Sadra:} {#1}} }

%
\IEEEpeerreviewmaketitle

\section{Introduction}
Videos contain visual information enriched by motion.
Motion is a useful cue for reasoning about human activities
or interactions between objects in a video. Given a few
initial frames of a video, our goal is to predict several frames into the future, as realistically as possible. By looking at a few frames, humans can predict what will happen next. Surprisingly, they can even attribute semantic meanings to random dots and recognize motion patterns~\cite{Johansson1973}. This shows
the importance of motion to infer the dynamics of the video
and to predict the future frames.

Motion cues have been heavily utilized for future frame
prediction in computer vision. A common approach is to factorize the video into static and dynamic components~\cite{Walker2015ICCV, Liu2017ICCV, Lu2017CVPR, Fan2019AAAI, Gao2019ICCV, Lotter2017ICLR, Brabandere2016NeurIPS, Vondrick2017CVPR}. First, most of the previous methods are deterministic and fail to model the uncertainty of the future. Second, motion is typically interpreted as local changes from one frame to the next. However, changes in
motion follow certain patterns when observed over some
time interval. Consider scenarios where objects move with
near-constant velocity, or humans repeating atomic actions
in videos. Regularities in motion can be very informative
for future frame prediction.

\begin{figure}[ht!]
    \centering
    \includegraphics[width=\linewidth]{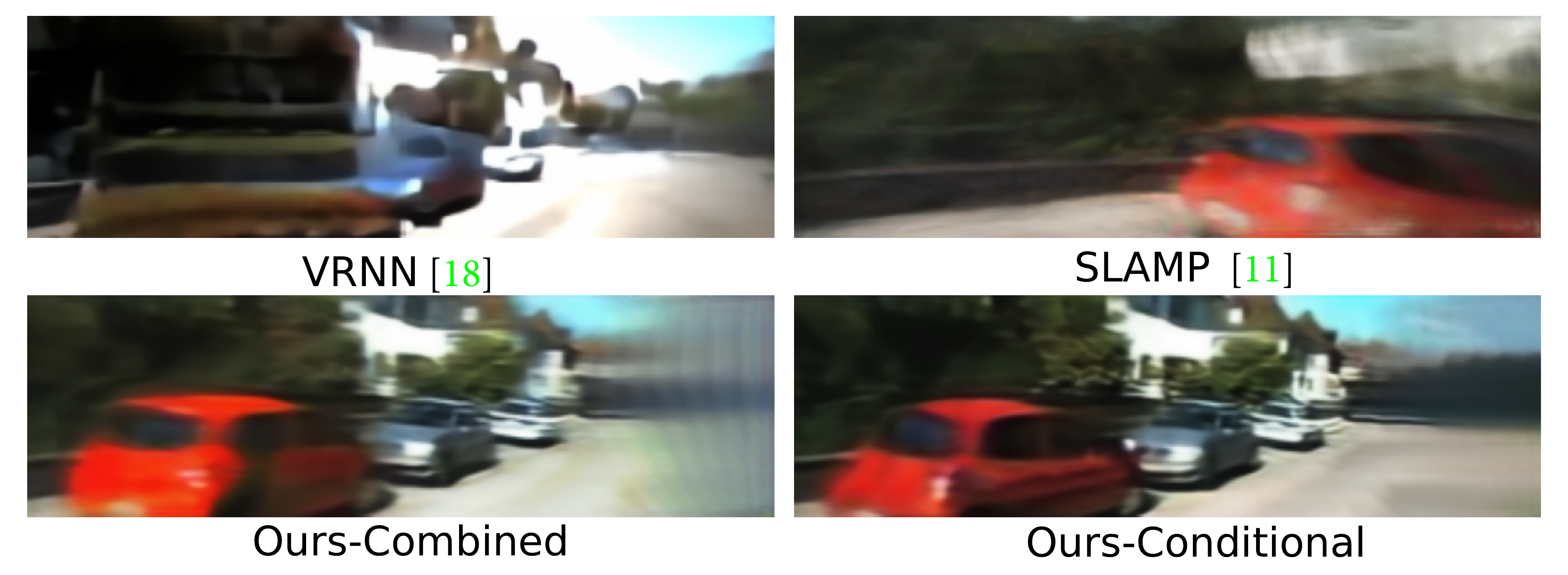}
    \caption{\textbf{Future Prediction while Turning.} We compare the future prediction results of previous methods (\textbf{top}) to ours (\textbf{bottom}) while the vehicle makes a right turn. While previous methods fail due to a less frequent scenario in the dataset, our method can generate a better future prediction due to explicit modeling of the structure and ego-motion. This figure shows a single frame after the conditioning frames, please see \figref{fig:qual_comp_kitti} for the whole sequence.}
    \label{fig:teaser}
\end{figure}

The world observed from a moving vehicle can be decomposed into a static part which moves only according to the motion of the vehicle, or the ego-motion, and a dynamic part containing independently moving objects. With two different types of motion, the future is quite uncertain and hard to predict but also crucial for the high-level decision making process of an autonomous vehicle.
The existing future frame prediction methods either ignore the uncertainty of the future or fail to fully capture the dynamics of the structured world by only focusing on changes in pixels. Despite drastic changes in the appearance of pixels, there are some common factors creating the observations which are shared across frames in a sequence. In this paper, we model, relate, and predict these factors to generate better predictions of future frames in a video. Reliable future predictions can help autonomous vehicle anticipate future and plan accordingly.

Observations from a moving vehicle depend on some factors which smoothly evolve through time. We exploit this continuity to predict future frames matching the observed sequences. 
We first factorize the underlying process leading to observations as the scene structure, the ego-motion of the vehicle, and the motion of the dynamic objects. After obtaining this factorization for the previous frames, we make future predictions for each component conditioned on the past.
Then, based on these predictions of the structure and the two types of motion, we generate future frames. In other words, rather than modelling the stochasticity of the future in the noisy pixel space, we model the stochasticity in terms of the underlying factors generating the pixels. In our experiments, we show that the structure and motion of the scene are continuous and can be propagated to the future more reliably than the pixels in real-world sequences.

\begin{figure*}[t!]
\centering
\includegraphics[width=\textwidth]{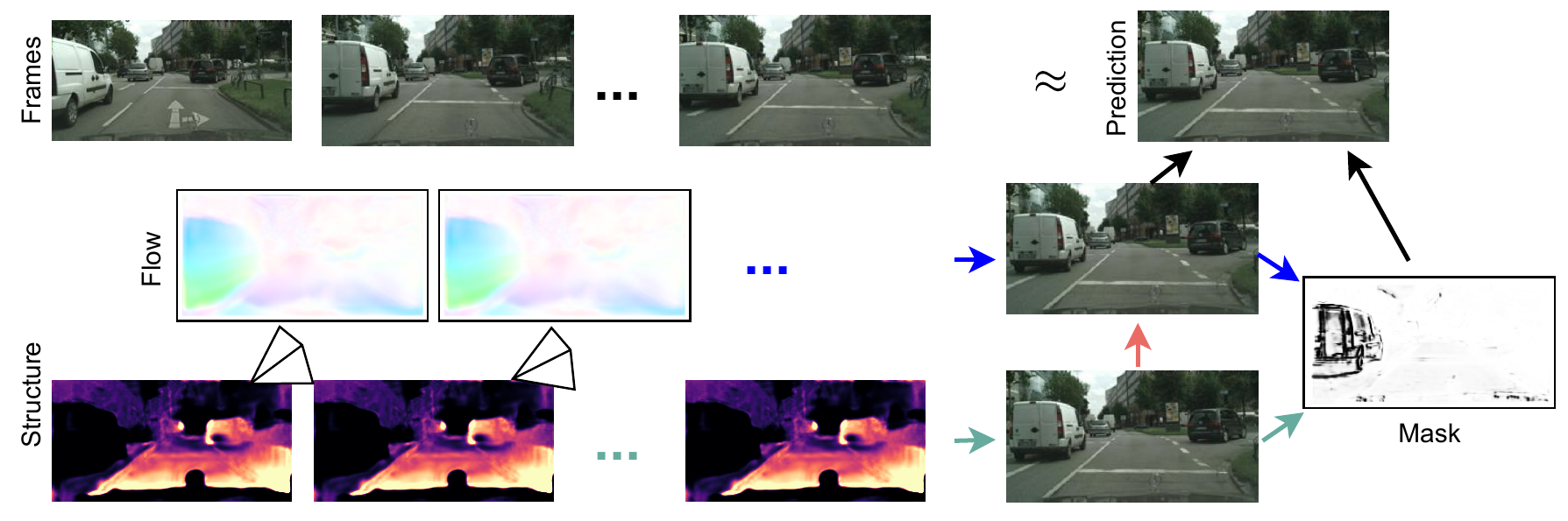}
\caption{\textbf{Overview.} For predicting future frames in a video~(\textbf{top}), we predict depth on each frame, and camera pose and optical flow between consecutive frames. We model the static part of the scene with depth and pose~(\textbf{bottom}). Conditioned on the static part, we model the remaining motion due to moving foreground objects with flow in the dynamic parts of the scene (\textbf{middle}). We obtain the final prediction~(\textbf{top-right}) by combining the static and the dynamic predictions based on the predicted mask~(\textbf{right}, dynamic in black). The moving car is missing in depth because it is predicted by the dynamic. The arrow from the static to the final prediction is omitted for clarity.}
\label{fig:overview}
\end{figure*}

The inherent uncertainty of the future has been addressed by the stochastic video prediction methods.
Earlier methods encode the dynamics of a video in stochastic latent variables which are then decoded to predict future frames~\cite{Denton2018ICML}. 
Our previous work~\cite{Akan2021ICCV} proposes to incorporate the motion history by explicitly predicting motion between consecutive frames. 
In particular, we learn two separate distributions representing the changes in the pixel space and the motion space. Similarly, in this work, we decompose the scene into static and dynamic components 
but we focus on driving scenarios where the static part also moves. By using the domain knowledge~\cite{Zhou2017CVPR, Godard2019ICCV}, we model the structure and the ego-motion together for the static part.
Then, we observe that the changes to the foreground objects are created by both the ego-motion and the independent motion of the object itself. In order to learn the object motion as residual motion on top of the ego-motion, we condition the dynamic latent variables on the static~\cite{Detlefsen2019NeurIPS}. 

To the best of our knowledge, our method is the first to decompose the motion in a scene by separating ego-motion and object motion for stochastic video prediction. We show that this separation improves the performance of future prediction in real-world scenes with a moving background and independently moving foreground objects on two real-world driving datasets, KITTI~\cite{Geiger2012CVPR, Geiger2013IJRR} and Cityscapes~\cite{Cordts2016CVPR}. Furthermore, conditioning the object motion on the ego-motion improves the results, especially for foreground objects in dynamic scenes of Cityscapes.

Our method performs on-par with the state-of-the-art method, Improved-VRNN~\cite{Castrejon2019ICCV} while being 40 $\times$ faster. Moreover, overall performance gaps compared to Improved-VRNN are due to background regions. Improved-VRNN performs better in the background regions, whereas our method’s performance in the foreground objects is better, which shows our model’s ability to capture dynamic objects. Our model is designed to predict future frames, but it can also generate future depth, pose, and optical flow to synthesize the target frame without even seeing the target frame. We evaluate our depth predictions in comparison to the state-of-the-art monocular depth estimation method~\cite{Godard2019ICCV}. Moreover, we evaluate our model in terms of diversity compared to Improved-VRNN. Our results show that our model can pinpoint the uncertainty into foreground regions or mostly moving objects whereas Improved-VRNN uniformly distributes the uncertainty over the whole scene. A preliminary, technical report version of our work is available online~\cite{Akan2022SLAMP3D}.

\section{Related Work}
\label{sec:rw}

\boldparagraph{Structure and Motion}
Our approach is related to view synthesis~\cite{Garg2016ECCV} where the target view can be synthesized by warping a source view based on the depth estimation from the target view~\cite{Jaderberg2015NeurIPS}. The difference between the synthesized target view and the original one can be used for self-supervised training.
Monocular depth estimation approaches~\cite{Zhou2017CVPR, Zhan2018CVPR, Wang2018CVPR, Bian2019NeurIPS, Mahjourian2018CVPR, Godard2019ICCV, Guizilini2020CVPR} generalize view synthesis to adjacent frames by also estimating the relative pose from one frame to the next. These methods can successfully model the structure and motion in the static part of the scene. However, the motion of independently moving objects remains as a source of error.
More recent works~\cite{Yin2018CVPR, Zou2018ECCV, Chen2019ICCV, Ranjan2019CVPR, Luo2019PAMI} estimate the residual optical flow to model the motion of independently moving objects. In addition, some of these methods model the consistency between depth and optical flow~\cite{Zou2018ECCV}, and also motion segmentation~\cite{Ranjan2019CVPR, Luo2019PAMI, Safadoust2021THREEDV}. In this paper, we similarly learn the decomposition of the world into static and dynamic parts but in longer sequences by exploiting the history from previous frames to predict future frames with uncertainty.

\boldparagraph{Stochastic Video Prediction}
SV2P~\cite{Babaeizadeh2018ICLR} and SVG~\cite{Denton2018ICML} are the first to model the stochasticity in video sequences using latent variables. The input from past frames are encoded in a posterior distribution to generate the future frames. In a stochastic framework, learning is performed by maximizing the likelihood of the observed data and minimizing the distance of the posterior distribution to a prior distribution, either fixed~\cite{Babaeizadeh2018ICLR} or learned from previous frames~\cite{Denton2018ICML}. Since time-variance is proven crucial by these previous works, we sample a latent variable at every time step. Sampled random variables are fed to a frame predictor, modelled recurrently using an LSTM. 
Typically, each distribution, including the prior and the posterior, is modeled with a recurrent model such as an LSTM.

There are various extensions to the basic formulation of stochastic video generation.
Villegas~\etal~\cite{Villegas2019NeurIPS} replace the linear LSTMs with convolutional ones at the cost of increasing the number of parameters. Castrejon~\etal~\cite{Castrejon2019ICCV} introduce a hierarchical representation to model latent variables at different scales which increases the complexity. Karapetyan~\etal~\cite{Karapetyan2022Arxiv} proposes usage of hierarchical recurrent networks to generate high quality video frames using multi-scale architectures.
Babaeizadeh~\etal~\cite{Babaeizadeh2021Arxiv} over-parameterize an existing architecture to first overfit to the training set, and then use data augmentation to generalize to the validation or test sets.
Lee~\etal~\cite{Lee2018ARXIV} incorporate an adversarial loss into the stochastic framework to generate sharper images, at the cost of less diverse results.
We also use convolutional LSTMs to generate diverse and sharp-looking results without any adversarial losses by first reducing the spatial resolution to reduce the cost. Future prediction is typically performed in the pixel space but there are other representations such as keypoints \cite{villegas2017ICML, Minderer2019NeurIPS} , coordinates \cite{gu2021densetnt, gao2020vectornet} and bird’s-eye view \cite{Hu2021ICCV, Akan2022Arxiv}.

\boldparagraph{Decomposition} State-space model SRVP~\cite{Franceschi2020ICML} learns a content variable from the first few frames which remains unchanged while predicting the future frames.
As shown in SLAMP~\cite{Akan2021ICCV}, the content variable in SRVP cannot handle changes in the background. In addition to pixel space, SLAMP separately models changes in motion and keeps track of a motion history. This reduces the role of the pixel decoder to recover occlusions, \eg around motion boundaries. 
We decompose the scene as static and dynamic where the static part not only considers the ego-motion but also the structure. This allows us to differentiate the motion in the background from the motion in the foreground which leads to a more meaningful scene decomposition for driving.
Disentanglement of explicit groups of factors has been explored before for generating 3D body models depending on the body pose and the shape~\cite{Detlefsen2019NeurIPS}. Inspired by the conditioning of the body pose on the shape, we condition the motion of the dynamic part on the static part to achieve a disentanglement between the two types of motion in driving.


\section{Methodology}
We investigate the effects of decomposing the scene into static and dynamic parts for stochastic video prediction, inspired by and built on top of SLAMP~\cite{Akan2021ICCV}.
We assume that the background or the static part of the scene, moves only according to the motion of the ego-vehicle and can be explained by the ego-motion and the scene structure.
We first predict future depth and ego-motion conditioned on a few given frames. This is different than SLAMP which predicts the static part in the pixel-space. By using the predictions of the structure and the ego-motion, we synthesize the static part of the scene. We model the remaining motion due to independently moving objects, the dynamic part of the scene, as the residual flow on top of the ego-motion. At the end, we combine the static and the dynamic predictions with a learned mask to generate the final predictions.

\subsection{Stochastic Video Prediction (SVP)}
Given the previous frames $\bx_{1:t-1}$ until time $t$, 
our goal is to predict the target frame $\bx_t$. For that purpose, we assume that we have access to the target frame $\bx_t$ during training and use it to capture the dynamics of the video in stochastic latent variables $\bz_t$. By learning to approximate the distribution over $\bz_t$, we can decode the future frame $\bx_t$ from $\bz_t$ and the previous frames $\bx_{1:t-1}$ at test time.

Using all the frames including the target frame, we compute a posterior distribution $q_{\bphi}\left(\bz_t \mid \bx_{1:t}\right)$ and sample a latent variable $\bz_t$ from this distribution at each time step. 
The stochastic process of the video is captured by the latent variable $\bz_t$. In other words, it should contain information accumulated over the previous frames rather than only condensing the information on the current frame. This is achieved by encouraging $q_{\bphi}\left(\bz_t \mid \bx_{1:t}\right)$ to be close to a prior distribution $p\left(\bz\right)$ in terms of KL-divergence. 
The prior can be sampled from a fixed Gaussian 
at each time step or can be learned from the previous frames up to the target frame $p_{\bpsi}\left(\bz_t \mid \bx_{1:t-1}\right)$. We prefer the latter as it is shown to work better by learning a prior that varies across time~\cite{Denton2018ICML}.

The target frame $\bx_t$ is predicted based on the information from the previous frames $\bx_{1:t-1}$ and the latent vectors $\bz_{1:t}$. In practice, we only use the latest frame $\bx_{t-1}$ and the latent vector $\bz_t$ as input and dependencies from further previous frames are propagated with a recurrent model. 
The output of the frame predictor $\bg_t$,
contains the information required to decode the target frame.
Typically, $\bg_t$ is decoded to a fixed-variance Gaussian distribution whose mean is the predicted target frame $\hat{\bx}_t$.

\subsection{SLAMP: SVP with Motion History}
In our previous work, we introduced SLAMP~\cite{Akan2021ICCV} by explicitly modeling the motion history to better predict the dynamic parts of the scene. We predict the target image, $\hat{\bx}_t$ not only in the pixel space as done in the previous work but also in the motion space. The pixel and motion space represent the static and the dynamic parts of the scene, respectively. More precisely, SLAMP simultaneously predicts the target frames in pixel space, $\bx_t^p$ which we call appearance prediction, and an optical flow, $\bff_{t-1:t}$ representing the motion from of the pixels from the previous time, $t-1$, to target time, $t$. By using the predicted optical flow, we reconstruct the target frame $\bx_t$ from the previous frame $\bx_{t-1}$ with differentiable warping \cite{Jaderberg2015NeurIPS}. Finally, SLAMP estimates a binary mask from two predictions to combine them as follows:
\begin{equation}
    \label{eq:slamp_mask}
    \hat{\bx}_t = \bm\left(\hat{\bx}_t^p, \hat{\bx}_t^f\right) \odot \hat{\bx}_t^p + \Big(\mathbf{1} - \bm\left(\hat{\bx}_t^p, \hat{\bx}_t^f\right)\Big) \odot \hat{\bx}_t^f
\end{equation}
where $\odot$ denotes element-wise Hadamard product and $\hat{\bx}_t^p$ is the result of appearance prediction and $\hat{\bx}_t^f$ is the result of warping the source frame to the target frame according to the estimated flow field, $\bff_{t-1:t}$.

The binary mask attends to the dynamic branch's prediction because in the dynamic parts of the scene, the target frame can be reconstructed by warping the previous frame using the motion information, \ie the predicted optical flow, $\bff_{t-1:t}$. However, motion prediction is unreliable in the occluded regions of the scene. SLAMP can recover the occluded pixels by attending to the appearance branch's prediction, $\bx_t^p$. The binary mask learns where to attend to in the appearance and the motion predictions by simply weighting them for the best possible target image reconstruction.

In order to disentangle the static and dynamic parts of the scene, SLAMP learns two different posterior distributions, $q_{\bphi_p}(\bz_t^p \mid \bx_{1:t})$ and $q_{\bphi_f}(\bz_t^f \mid \bx_{1:t})$, respectively. By sampling two different latent variables, $\bz^p_t$ and $\bz^f_t$, from these posterior distributions, SLAMP achieves the disentanglement of static and dynamic parts of the scene. We expect the dynamic component to focus on changes, whereas the static component to focus on what remains constant from the previous frames to the target frame. If the background moves in response to camera movement, the static component can simulate the change in the background as long as it remains consistent during the video, \eg the ego-motion of a car.

\boldparagraph{Motion history}
Instead of modelling local temporal changes, \ie from last frame to the target frame, we accumulate the past motion information in the previous frames into the latent variable  $\bz^f_t$. To achieve this, we learn a posterior distribution, $q_{\bphi_f}(\bz_t^f \mid \bx_{1:t})$ as close as possible to a prior distribution by minimizing the KL-divergence. Following \cite{Denton2018ICML}, we learn the motion prior distribution from the past frames, $p_{\bpsi_f}(\bz_t^f \mid \bx_{1:t-1})$. This way, we accumulate the past motion information in the scene into the prior distribution.  We repeat this process for the static part $\bz_t^p$ with posterior $q_{\bphi_p}(\bz_t^p \mid \bx_{1:t})$ and the learned prior $p_{\bpsi_p}(\bz_t^p \mid \bx_{1:t-1})$ to model the changes in the static parts of the scene.

\subsection{From SLAMP to SLAMP-3D}
In this work, we propose to extend SLAMP~\cite{Akan2021ICCV} to SLAMP-3D by incorporating scene structure into the estimation. Similar to SLAMP, we decompose the scene into static and dynamic components where the static component focus on the changes in the background due to camera motion and the dynamic component on the remaining motion in the scene due to independently moving objects. There are two major differences with respect to SLAMP~\cite{Akan2021ICCV}. First, in this work, we model the static parts of the scene in the motion space as well to better represent the changes in the background due to the ego-motion of the vehicle.
Second, we condition the prediction of dynamic component on the static to predict the residual motion in the scene, \ie the remaining motion after the scene moves according to ego-motion due to the independently moving objects. 

\begin{figure}[t!]
\centering
\includegraphics[width=\linewidth]{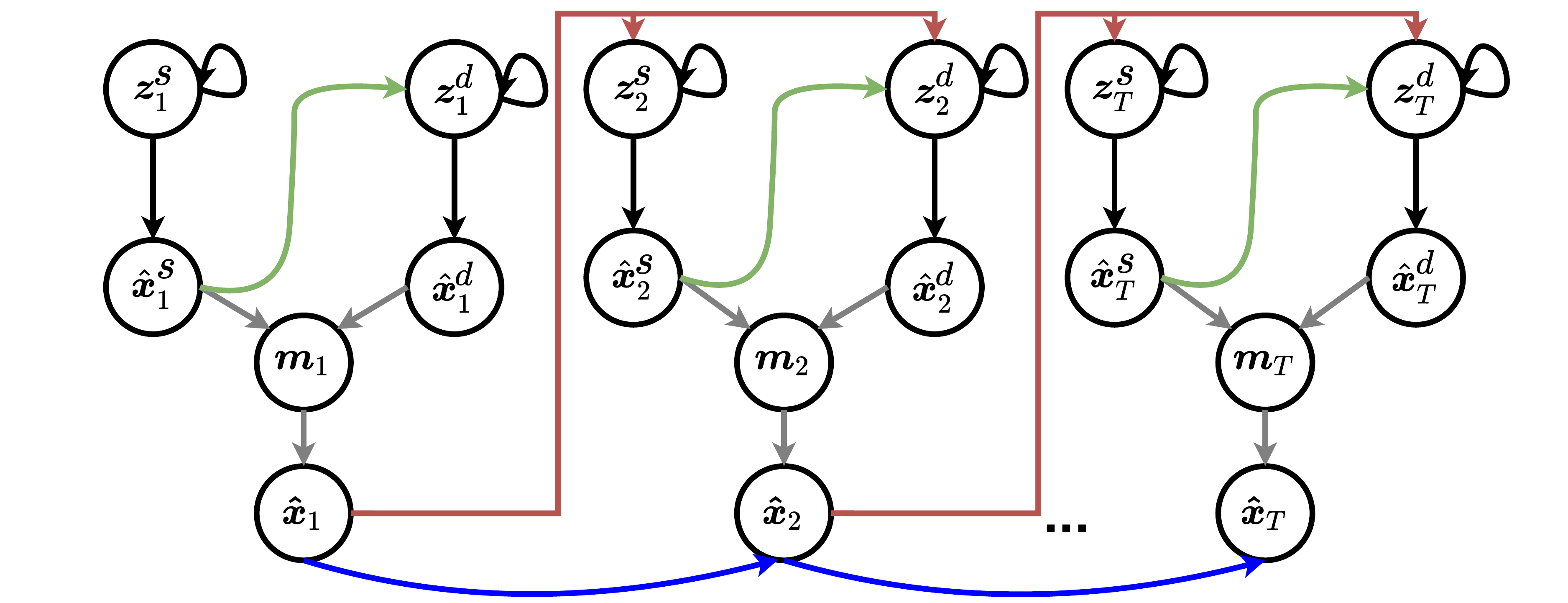}
\caption{\textbf{Graphical Model.} The graphical model shows the generation process of our model with static and dynamic latent variables $\bz^s_{t}$ and $\bz^d_{t}$ generating frames $\hat{\bx}^s_{t}$ and $\hat{\bx}^d_{t}$.
Information is propagated between time-steps through the recurrence between frame predictions (\textcolor{blue}{blue}), corresponding latent variables, and from frame predictions to latent variables (\textcolor{OrangeRed}{red}). The dependency of dynamic latent variables on static is shown in \textcolor{LimeGreen}{green}.}
\label{fig:graphical}
\end{figure}

Similar to SLAMP, we compute two separate distributions, $q_{\bphi_s}\left(\bz_t^s \mid \bx_{1:t}\right)$ and $q_{\bphi_d}\left(\bz_t^d \mid \bx_{1:t}\right)$, for static and dynamic components, respectively. This way, both distributions focus on different parts of the scene which decomposes scene into static and dynamic parts. Since the camera motion applies both to the background and to the objects, dynamic component only needs to learn the residual motion. We achieve this by introducing an explicit conditioning of the dynamic latent variable $\bz_t^d$ on the static latent variable $\bz_t^s$:
\begin{align}
    \label{eq:slamp3d_prior_post}
    p\left(\bz_t^s \mid \bx_{1:t}\right) &\approx q_{\bphi_s}\left(\bz_t^s \mid \bx_{1:t}\right) \quad \text{and} \\
    p\left(\bz_t^d \mid \bx_{1:t}\right) &\approx q_{\bphi_d}\left(\bz_t^d \mid \bx_{1:t}, \bz_t^s\right) \nonumber
\end{align}
With two latent variables, we extend the stochasticity to the motion space, separately for the ego-motion and the object motion. Furthermore, we condition the dynamic component on the static to define the object motion as residual motion that remains after explaining the scene according to camera motion, similar to the disentanglement of the body pose and the shape in \cite{Detlefsen2019NeurIPS}.

\boldparagraph{Two Types of Motion History} Similar to SLAMP, we model the motion history in SLAMP-3D but by disentangling it to ego-motion and residual motion. Essentially, we learn two separate motion histories for the static and the dynamic components. The latent variables $\bz_t^s$ and $\bz_t^d$ contain motion information accumulated over the previous frames rather than local temporal changes between the last frame and the target frame. This is achieved by encouraging each posterior, $q_{\bphi_s}\left(\bz_t^s \mid \bx_{1:t}\right)$ and $q_{\bphi_d}\left(\bz_t^d \mid \bx_{1:t}, \bz_t^s\right)$, to be close to a prior distribution in terms of KL-divergence. 
Similar to SVG~\cite{Denton2018ICML} and SLAMP~\cite{Akan2021ICCV}, we learn each prior distribution from the previous frames up to the target frame, $p_{\bpsi_s}\left(\bz_t^s \mid \bx_{1:t-1}\right)$ and $p_{\bpsi_d}\left(\bz_t^d \mid \bx_{1:t-1}\right)$. Note that we learn separate posterior and prior distributions for the static and the dynamic components. The static contains depth and pose encoding, while the dynamic contains the flow encoding conditioned on the static as explained next.

\subsection{SLAMP-3D: SVP with Structure and Motion}
\begin{figure*}[t!]
     \begin{subfigure}{.48\linewidth}
        \includegraphics[width=\linewidth]{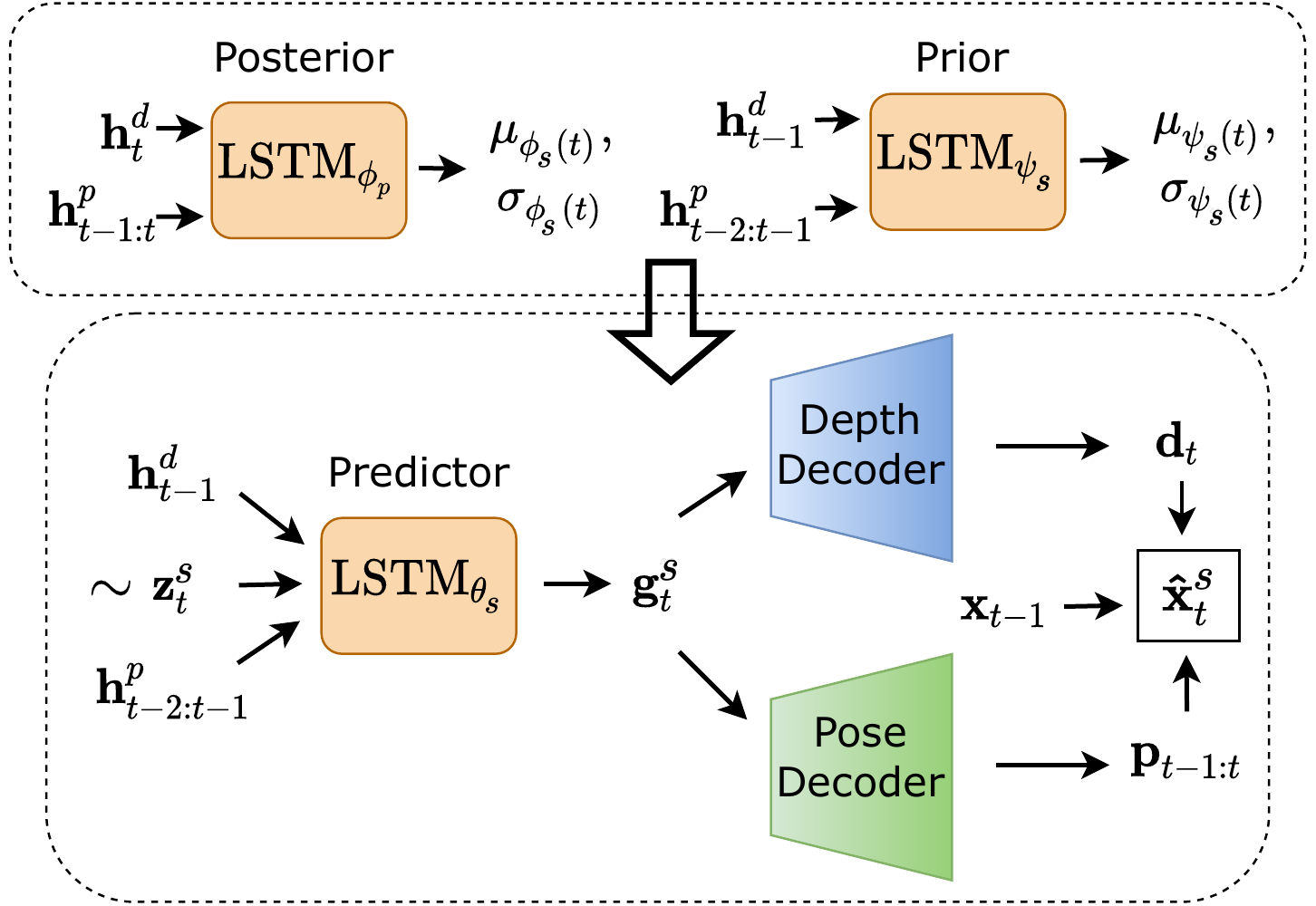}
        \caption{Static Architecture}
        \label{fig:arch_static}
    \end{subfigure}
    \begin{subfigure}{.48\linewidth}
        \includegraphics[width=\linewidth]{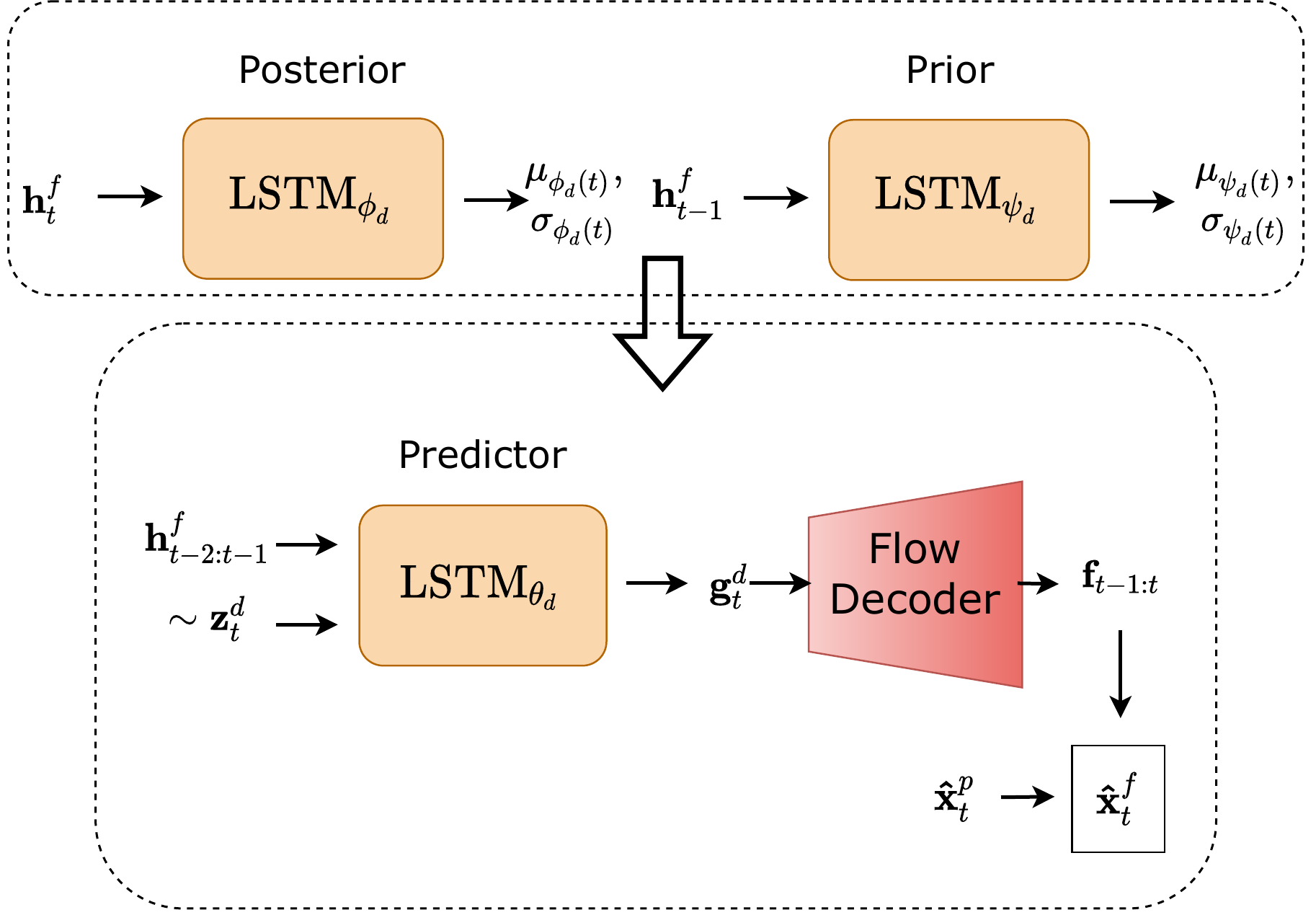}
        \caption{Dynamic Architecture}
        \label{fig:arch_dynamic}
    \end{subfigure}
\caption{{\bf Architecture.} 
The architecture in (\subref{fig:arch_static}) shows the sampling of the static latent variable $\bz_t^s$ from the posterior $\bphi_s$ or the prior $\bpsi_s$, which is then used to predict $\bg_t^s$, the future representation of the static part. The future depth $\bd_t$ and pose $\bp_{t-1:t}$ are decoded and used to generate the static prediction $\hat{\bx}_t^s$. The architecture in (\subref{fig:arch_dynamic}) shows the similar procedure for the dynamic part using optical flow with the flow decoder instead of depth and pose.}
\label{fig:model_details}
\vspace{-0.65cm}
\end{figure*}
Inspired by the previous work on unsupervised learning of depth and ego-motion~\cite{Zhou2017CVPR, Godard2019ICCV, Casser2019AAAI, Guizilini2020CVPR, Safadoust2021THREEDV}, flow~\cite{Jason2016ECCV, Ren2017AAAI, Meister2018AAAI}, as well as the relating of the two~\cite{Zou2018ECCV, Yin2018CVPR, Ranjan2019CVPR, Chen2019ICCV}, we reconstruct the target frame from the static and the dynamic components at each time-step.

\boldparagraph{Depth and Pose} For the static component which moves only according to the camera motion, we estimate the relative camera motion or the camera pose $\bp_{t-1:t}$ from the previous frame $t-1$ to the target frame $t$. The pose $\bp_{t-1:t}$ represents the 6-degrees of freedom rigid motion of the camera from the previous frame to the target frame. We also estimate the depth of the target frame, $\bd_{t}$, and reconstruct the target frame as $\hat{\bx}_t^s$ from the previous frame $\bx_{t-1}$ using the estimated pose $\bp_{t-1:t}$ and depth $\bd_{t}$ via differentiable warping~\cite{Jaderberg2015NeurIPS}. Note that we predict the depth and pose a priori, without actually seeing the target frame in contrast to the previous unsupervised monocular depth approaches that use the target frame to predict depth and pose~\cite{Zhou2017CVPR, Godard2019ICCV, Casser2019AAAI, Guizilini2020CVPR}.

\boldparagraph{Residual Flow} We estimate optical flow $\bff_{t-1:t}$ as the remaining motion from the reconstruction of the target frame $\hat{\bx}_t^s$ to the target frame $\bx_t$. The flow $\bff_{t-1:t}$ represents the motion of the pixels belonging to objects that move independently from the previous frame to the target frame. We reconstruct the target frame as $\hat{\bx}_t^d$ from the static prediction, $\hat{\bx}^s_t$, by using the estimated optical flow $\bff_{t-1:t}$.
Explicit conditioning of the optical flow on the static prediction allows the network to learn the remaining motion in the scene from the source frame to the target frame which corresponds to the motion that cannot be explained by the camera pose. As in the case of depth and pose prediction, we predict the flow from the reference frame to the target frame without actually seeing the target frame. In other words, we predict future motion based on the motion history, which is different than regular optical flow approaches that use the target frame for prediction~\cite{Jason2016ECCV, Ren2017AAAI, Meister2018AAAI}.

Note that the role of optical flow in SLAMP-3D is different than SLAMP~\cite{Akan2021ICCV}. In SLAMP-3D, the optical flow is predicted conditioned on the static component. This way, the optical flow in SLAMP-3D corresponds to the residual flow as opposed to full flow as in the case of SLAMP. Therefore, we warp the \emph{static component's prediction} instead of the previous frame as in the case of SLAMP.

\boldparagraph{Combining Static and Dynamic Predictions} Similar to SLAMP~\cite{Akan2021ICCV}, we use the Hadamard product to combine the static prediction $\hat{\bx}_t^s$ and the dynamic prediction $\hat{\bx}_t^d$ into the final prediction $\hat{\bx}_t$. The same equation \eqref{eq:slamp_mask} applies by changing $\hat{\bx}_t^p$ and $\hat{\bx}_t^f$ to $\hat{\bx}_t^s$ and $\hat{\bx}_t^d$, respectively.
The static parts in the background can be reconstructed accurately using the depth and camera pose estimation. For the dynamic parts with moving objects, the target frame can be reconstructed accurately using the residual motion prediction.
The mask prediction learns a weighting between the static and the dynamic predictions for combining them into the final prediction. 

\subsection{Variational Inference}
\label{sec:inf}
Removing time dependence for clarity, \eqnref{eq:cond_prob} expresses the conditional joint probability corresponding to the graphical model shown in \figref{fig:graphical}:
\begin{equation}
    \label{eq:cond_prob}
    p_{\btheta}(\bx) = \iint p(\bx \mid \bz^s, \bz^d)~p(\bz^s)~p(\bz^d) ~\mathrm{d}\bz^s ~\mathrm{d}\bz^d
\end{equation}
SLAMP's variational inference can be obtained by simply changing the latent variables to $\bz^p$ and $\bz^f$.
The true distribution over the latent variables $\bz_{t}^{s}$ and $\bz_{t}^{d}$ is intractable. We assume that the dynamic component $\bz_{t}^{d}$ depends on the static component $\bz_{t}^{s}$ and train time-dependent inference networks $q_{\bphi_s}\left(\bz_{t}^{s} \mid \bx_{1:T}\right)$ and $q_{\bphi_d}\left(\bz_{t}^{d} \mid \bx_{1:T}, \bz_{t}^{s}\right)$ to approximate the true distribution with conditional Gaussian distributions. 
In order to optimize the likelihood of $p\left(\bx_{1:T}\right)$, we need to infer latent variables $\bz_{t}^{s}$ and $\bz_{t}^{d}$, which correspond to uncertainty in static and dynamic parts of future frames, respectively. We use a variational inference model to infer the latent variables.

At each time-step, $\bz_{t}^{d}$ depends on $\bz_{t}^s$ but each is independent across time. Therefore, we can decompose Kullback-Leibler terms into individual time steps. We train the model by optimizing the variational lower bound as shown in \eqref{eq:elbo} where each $\bz_{1:t}^\cdot$ is sampled from the respective posterior distribution $q_{\bphi_{\cdot}}$~(see Appendix for the derivation). By changing the respective latent variables into $\bz^p$ and $\bz^f$ and losing the conditioning of dynamic component, we can derive SLAMP's evidence lower bound. 

\begin{align}
\label{eq:elbo}
\log p_{\btheta}(\bx) \geq 
&\mathbb{E}_{q_s(\bz^s \mid \bx)} \left[ \mathbb{E}_{q_d(\bz^d \mid \bz^s, \bx)}\left[ \log p(\bx \mid \bz^s, \bz^d)\right] \right] - \\ 
&\mathbb{E}_{q_s(\bz^s \mid \bx)} \left[D_{KL}(q_d(\bz^d \mid \bz^s, \bx) \mid\mid p(\bz^d)) \right] - \nonumber \\
&D_{KL}(q_s(\bz^s \mid \bx) \mid\mid p(\bz^s)) \nonumber
\end{align}

The likelihood $p_{\btheta}$ can be interpreted as minimizing the difference between the actual frame $\bx_t$ and the prediction $\hat{\bx}_t$ as defined in \eqnref{eq:slamp_mask} by changing $p$ and $f$ to $s$ and $d$. We apply the reconstruction loss to the predictions of static and dynamic components as well.
The posterior terms for uncertainty are estimated as an expectation over $q_{\bphi_s}\left(\bz_t^s \mid \bx_{1:t}\right)$ and $q_{\bphi_d}\left(\bz_t^d \mid \bx_{1:t}, \bz_t^s\right)$.
For the second term, we use sampling to approximate the expectation as proposed in \cite{Detlefsen2019NeurIPS}.
As in \cite{Denton2018ICML}, we also learn the prior distributions from the previous frames up to the target frame as 
$p_{\bpsi_s}\left(\bz_t^s \mid \bx_{1:t-1}\right)$ and $p_{\bpsi_d}\left(\bz_t^d \mid \bx_{1:t-1}\right)$.
We train the model using the re-parameterization trick \cite{Kingma2014ICLR} and choose the posteriors to be factorized Gaussian so that all the KL divergences can be computed analytically.
We apply the optimal variance estimate as proposed in $\sigma$-VAE~\cite{Rybkin2021ICML} to learn an optimal value for the hyper-parameter corresponding to the weight of the KL term. 

\subsection{Architecture}
\label{sec:arch}
We encode the frames with a feed-forward convolutional architecture to obtain frame-wise features at each time-step. Our goal is to reduce the spatial resolution of the frames in order to reduce the complexity of the model.  
On top of this shared representation, we have a depth head and a pose head to learn an encoding of depth at each frame, $\bh_{t}^d$, and an encoding of pose for each consecutive frame pair, $\bh_{t-1:t}^p$.
As shown in \figref{fig:model_details}, we train convolutional LSTMs to infer the static distribution at each time-step from the encoding of depth and pose:
\begin{align}
    \label{eq:static_dist}
    &\be_{t-1},~\be_t = \mathrm{ImgEnc}\left(\bx_{t-1}\right),~ \mathrm{ImgEnc}\left(\bx_t\right) \\
    &\bh_t^d,~\bh_{t-1:t}^p = \mathrm{DepthEnc}\left(\be_t\right),~ ~\mathrm{PoseEnc}\left(\be_{t-1}, \be_t\right) \nonumber \\
    &\bz_t^s \sim \mathcal{N}\left(\bmu_{\bphi_s(t)}, \bsigma_{\bphi_s(t)}\right) \quad \text{where} \nonumber \\ &\bmu_{\bphi_s(t)},~ \bsigma_{\bphi_s(t)} = \mathrm{LSTM}_{\bphi_s}\left(\bh_t^d,~ \bh_{t-1:t}^p\right) \nonumber
     \nonumber
\end{align}
\eqnref{eq:static_dist} only shows the posterior distribution, similar steps are followed for the static prior by using the depth and pose representations from the previous time-step.

At the first time-step where there is no previous pose estimation, we assume zero-motion by estimating the pose from the previous frame to itself.
We learn a static frame predictor to generate the target frame based on the encoding of depth and pose and the static latent variable $\bz_t^s$: 
\begin{equation}
    \label{eq:static_frame_pred}
    \bg_t^s = \mathrm{LSTM}_{\btheta_s}\left(\bh_{t-1}^d, \bh_{t-2:t-1}^p, \bz_t^s\right)
\end{equation}
Then, the depth and pose estimations are decoded from $\bg_t^s$. 
Based on the depth and pose estimations, the static prediction $\hat{\bx}_t^s$ is reconstructed by inverse warping the previous frame $\bx_{t-1}$. In SLAMP, we encode only the pixel information instead of depth and pose. We can obtain SLAMP's architecture by replacing depth and pose with pixel to model the appearance only without the structure and ego-motion.

We learn an encoding of the remaining motion, $\bh_{t}^f$, in the dynamic parts from the output of the static frame predictor $\bg_{t}^s$ to the encoding of the target frame $\bx_t$.
We train convolutional LSTMs to infer dynamic posterior and prior distributions at each time-step from the encoded residual motion.
The posterior LSTM is updated based on the $\bh_{t}^f$ and the latent variable $\bz_t^d$ is sampled from the posterior:
\begin{align}
    &\bh_t^f = \mathrm{MotionEnc}\left(\bg_t^s, \be_t\right) \\
    &\bz_t^d \sim \mathcal{N}\left(\bmu_{\bphi_d(t)}, \bsigma_{\bphi_d(t)}\right) \quad \text{where} \nonumber \\ &\bmu_{\bphi_d(t)},~ \bsigma_{\bphi_d(t)} = \mathrm{LSTM}_{\bphi_d}\left(\bh_t^f\right) \nonumber
\end{align}

For the dynamic prior, we use the motion representation from the previous time step to update the prior LSTM and sample the $\bz_t^d$ from it.
Similar to the pose above, we assume zero-motion at the first time-step where there is no previous motion.
The dynamic predictor LSTM is updated according to encoded features and sampled latent variables:
\begin{equation}
    \bg_t^d = \mathrm{LSTM}_{\btheta_d}\left(\bh_{t-1}^f, \bz_t^d\right)
\end{equation}
%
During training, the latent variables are sampled from the posterior distribution. At test time, they are sampled from the posterior for the conditioning frames and from the prior for the following frames. By removing explicit conditioning of the dynamic component on the static, we can obtain the SLAMP's motion component.

After we generate target time's features for static and dynamic components, $\bg_t^s$ and $\bg_t^d$, respectively, we decode them into depth and pose for the static component, optical flow for the dynamic component. We first predict the depth values of the target frame without seeing the frame and the pose from the previous frame to the target frame using the static features, $\bg_t^s$. Then, we warp the previous frame $\bx_{t-1}$ using the predicted depth and pose to obtain the static prediction for the target frame, $\hat{\bx}_t^s$. For the dynamic part, we predict the residual flow from the dynamic features, $\bg_t^d$. Then, we warp the static frame prediction $\hat{\bx}_t^s$, using the predicted residual flow to obtain the dynamic prediction, $\hat{\bx}_t^d$. Finally, static and dynamic predictions are combined into the final prediction, $\hat{\bx}_t$, using the predicted mask as shown in \eqref{eq:slamp_mask}.

\section{Experiments}
\begin{table*}[t!]
    \caption{
        \label{tab:results}
        \textbf{Quantitative Results.} We compare our \emph{Combined} and \emph{Conditional} models to the other video prediction approaches. The best results are shown in bold and the second best underlined.
    }
    \sisetup{detect-weight, table-align-uncertainty=true, mode=text}
    \renewrobustcmd{\bfseries}{\fontseries{b}\selectfont}
    \renewrobustcmd{\boldmath}{}
    \centering
    \resizebox{\textwidth}{!}{\begin{tabular}{lccc|ccc}%
        \toprule
        \multirow{2}{*}{Models} &  \multicolumn{3}{c}{KITTI \cite{Geiger2012CVPR, Geiger2013IJRR}} & \multicolumn{3}{c}{Cityscapes \cite{Cordts2016CVPR}} \\ \cmidrule{2-7}
        {} & PSNR ($\uparrow$) & SSIM ($\uparrow$) & LPIPS ($\downarrow$) & PSNR ($\uparrow$) & SSIM ($\uparrow$) & LPIPS ($\downarrow$) \tabularnewline
        \midrule
        SVG~\cite{Denton2018ICML} & 12.70 $\pm$ 0.70 & 0.329 $\pm$ 0.030 &  0.594 $\pm$ 0.034 & 20.42 $\pm$ 0.63 & 0.606 $\pm$ 0.023 & 0.340 $\pm$ 0.022 \tabularnewline
        SRVP~\cite{Franceschi2020ICML}  & 13.41 $\pm$ 0.42 & 0.336 $\pm$ 0.034 & 0.635 $\pm$ 0.021 & 20.97 $\pm$ 0.43 & 0.603 $\pm$ 0.016 & 0.447 $\pm$ 0.014 \tabularnewline
        SLAMP~\cite{Akan2021ICCV} & 13.46 $\pm$ 0.74 & 0.337 $\pm$ 0.034 & 0.537 $\pm$ 0.042 & \bfseries 21.73 $\pm$ 0.76 & \bfseries 0.649 $\pm$ 0.025 & \underline{0.294} $\pm$ 0.022 \tabularnewline
        Improved-VRNN~\cite{Castrejon2019ICCV} &  14.15 $\pm$ 0.47 & \underline{0.379} $\pm$ 0.023 & \bfseries 0.372 $\pm$ 0.020 & 21.42 $\pm$ 0.67 &  {0.618} $\pm$ 0.020 & \bfseries 0.260 $\pm$ 0.014 \tabularnewline
        Ours-Combined & \bfseries 14.45 $\pm$ 0.35 &  0.378 $\pm$ 0.019 &  0.533 $\pm$ 0.016 & 21.00 $\pm$ 0.41 &  0.631 $\pm$ 0.014 & 0.309 $\pm$ 0.009 \tabularnewline
        Ours-Conditional &  \underline{14.32} $\pm$ 0.33 & \bfseries 0.383 $\pm$ 0.020 &  0.501 $\pm$ 0.016 &  \underline{21.43} $\pm$ 0.43 & \underline{0.643} $\pm$ 0.014 & {0.306} $\pm$ 0.009 \tabularnewline
        \bottomrule
    \end{tabular}}
    \vspace{-3mm}
\end{table*}

\boldparagraph{Implementation Details}
We first process each image with a shared backbone~\cite{Simonyan2015ICLR, He2016CVPR}
to reduce the spatial resolution.
We add separate heads for extracting features for depth, pose, and flow. We learn the static distribution based on depth and pose features, and dynamic distribution based on flow features with two separate ConvLSTMs. Based on the corresponding latent variables, we learn to predict the next frame's static and dynamic representation with another pair of ConvLSTMs. 
We decode depth and pose from the static representation and flow from the dynamic. 
We obtain the static frame prediction by warping the previous frame according to depth and pose, and dynamic prediction by warping the static prediction according to flow. We train the model using the negative log likelihood loss. 

\boldparagraph{Datasets} 
We perform experiments on two challenging autonomous driving datasets, KITTI~\cite{Geiger2012CVPR, Geiger2013IJRR} and Cityscapes~\cite{Cordts2016CVPR}. 
Both datasets contain everyday real-world scenes with complex dynamics due to both background and foreground motion. 
We train our model on the training set of the Eigen split on KITTI~\cite{Eigen2014NeurIPS}. We apply the pre-processing followed by monocular depth approaches and remove the static scenes~\cite{Zhou2017CVPR}. In addition, we use the depth ground-truth on KITTI to validate our depth predictions.
Cityscapes primarily focuses on semantic understanding of urban street scenes, therefore contains a larger number of dynamic foreground objects compared to KITTI. However, motion lengths are larger on KITTI due to lower frame-rate. 
On both datasets, we condition on 10 frames and predict 10 frames into the future to train our models. Then, at test time, we predict 20 frames conditioned on 10 frames.

\boldparagraph{Evaluation Metrics}
We compare the video prediction performance using four different metrics:
Peak Signal-to-Noise Ratio (PSNR) based on the $L_2$ distance between the frames penalizes differences in dynamics but also favors blurry predictions.
Structured Similarity (SSIM) compares local patches to measure similarity in structure spatially. %
Learned Perceptual Image Patch Similarity (LPIPS)~\cite{Zhang2018CVPR} measures the distance between learned features extracted by a CNN trained for image classification.
Frechet Video Distance (FVD) \cite{Unterthiner2019ARXIV}, lower better, compares temporal dynamics of generated videos to the ground truth in terms of representations computed for action recognition.

\begin{table}[h!]
    \sisetup{detect-weight, table-align-uncertainty=true, mode=text}
    \renewrobustcmd{\bfseries}{\fontseries{b}\selectfont}
    \renewrobustcmd{\boldmath}{}
    \centering
    \caption{
        \textbf{Baselines.} Results of our method~(\emph{Conditional}) by removing the dynamic part~(\emph{Depth-Only}) and by removing the conditioning of the dynamic part on the static~(\emph{Combined}).}
    \label{tab:ablation}
\begin{subtable}[h]{\linewidth}
    \resizebox{\linewidth}{!}{\begin{tabular}{lccc}
        \toprule
        Models &  PSNR ($\uparrow$) & SSIM ($\uparrow$) & LPIPS ($\downarrow$) \tabularnewline 
        \midrule
        Depth-Only & 13.02 $\pm$ 0.44 & 0.301 $\pm$ 0.021 & 0.523 $\pm$ 0.014 
        \tabularnewline
        Combined  & \bfseries 14.45 $\pm$ 0.35 &   0.378 $\pm$ 0.019 & 0.533 $\pm$ 0.016 \tabularnewline
        Conditional &  14.32 $\pm$ 0.33 & \bfseries 0.383 $\pm$ 0.020 & \bfseries 0.501 $\pm$ 0.016
        \tabularnewline
        \bottomrule
    \end{tabular}
    }
    \vspace{1mm}
    \subcaption{KITTI~\cite{Geiger2012CVPR, Geiger2013IJRR}}
    \vspace{2mm}
\end{subtable}
\begin{subtable}[h]{\linewidth}
    \resizebox{\linewidth}{!}{\begin{tabular}{lccc}
        \toprule
        Models &  PSNR ($\uparrow$) & SSIM ($\uparrow$) & LPIPS ($\downarrow$) \tabularnewline 
        \midrule
        Depth-Only & 19.97 $\pm$ 0.48 & 0.580 $\pm$ 0.016 & 0.445 $\pm$ 0.014
        \tabularnewline
        Combined  & 21.00 $\pm$ 0.41 & 0.631 $\pm$ 0.014 & 0.309 $\pm$ 0.009 \tabularnewline
        Conditional & \bfseries 21.43 $\pm$ 0.43 & \bfseries 0.643 $\pm$ 0.014 & \bfseries 0.306 $\pm$ 0.009
        \tabularnewline       \bottomrule
    \end{tabular}
    }
    \vspace{1mm}
    \subcaption{Cityscapes~\cite{Cordts2016CVPR}}
\end{subtable}
\vspace{-0.6cm}
\end{table}

\subsection{Ablation Study}
We first examine the importance of each contribution on KITTI and Cityscapes in \tabref{tab:ablation}. We start with a simplified version of our model called \emph{Depth-Only} by removing the dynamic latent variables and the flow decoder. This model inherits the weakness of depth and ego-motion estimation methods by assuming a completely static scene. Therefore, it cannot model the motion of the dynamic objects but it still performs reasonably well, especially on KITTI, since the background typically covers a large portion of the image. In the next two models, we include dynamic latent variables. 
We evaluate the importance of conditioning of dynamic variables on the static, \emph{Combined} versus \emph{Conditional}. For the \emph{Combined}, we independently model static and dynamic latent variables and simply combine them in the end with the predicted mask. For the \emph{Conditional}, we conditioned the dynamic latent variable on the static latent variable. First, both models improve the results compared to the \emph{Depth-Only} case. This confirms our intuition about modelling the two types of motion in the scene separately. The two perform similarly on KITTI but the \emph{Conditional} outperforms the \emph{Combined} on Cityscapes due to larger number of moving foreground objects.

\subsection{Quantitative Results}
We trained and evaluated stochastic video prediction methods on KITTI and Cityscapes by using the same number of conditioning frames including SVG~\cite{Denton2018ICML}, SRVP~\cite{Franceschi2020ICML}, SLAMP~\cite{Akan2021ICCV}, and Improved-VRNN~\cite{Castrejon2019ICCV}. We optimized their architectures for a fair comparison~(see Appendix).

\boldparagraph{Frame-Level Evaluations}
Recent methods including SLAMP~\cite{Akan2021ICCV}, Improved-VRNN~\cite{Castrejon2019ICCV}, and our models clearly outperform SVG~\cite{Denton2018ICML} and SRVP~\cite{Franceschi2020ICML} in terms of all three metrics. Improved-VRNN \cite{Castrejon2019ICCV} \footnote{The results of Improved-VRNN on Cityscapes is different from the ones in the original paper since we retrained their model on a similar resolution to ours with the same number of conditioning frames as ours.} achieves that by using five times more parameters compared to our models~(57M vs. 308M). A recent study shows the importance of attacking the under-fitting issue in video prediction~\cite{Babaeizadeh2021Arxiv}. The results can be improved by over-parameterizing the model and using data augmentation to prevent over-fitting. This finding
is complementary to other approaches including ours, however, it comes at a cost in terms of run-time. The time required to generate 10 samples is 40~seconds for Improved-RNN compared to 1~second, or significantly less in case of SRVP, for other methods. %

SLAMP~\cite{Akan2021ICCV} achieves the top-performing results on Cityscapes in terms of PSNR and SSIM by explicitly modelling the motion history. The separation between the pixel and the motion space is achieved by learning a separate distribution for each with a slight increase in complexity compared to SVG. We follow a similar decomposition as static and dynamic but we differentiate between the two types of motion in driving. As a result, our models clearly outperform SLAMP on KITTI where the camera motion is large due to low frame-rates. In addition, our models can produce reliable depth predictions for the static part of the scene (\tabref{tab:depth_eval}).
In SLAMP, we report results on generic video prediction datasets such as MNIST, KTH~\cite{Schuldt2004CVPR} and BAIR~\cite{EbertFLL17} as well. 
In SLAMP-3D, we focus on driving scenarios by utilizing the domain knowledge for a better modelling of the static part. In case of SLAMP, the pixel decoder only focuses on cases that cannot be handled by the flow decoder, \eg occlusions. Whereas in our case, the static models the whole background, leading not only to a better segmentation of the background in the mask but also to better results in the background on KITTI, as shown in \tabref{tab:fg_bg_results_kitti}.

\begin{table}[h!]
    \captionsetup[table]{justification=centering}
    \renewrobustcmd{\bfseries}{\fontseries{b}\selectfont}
    \renewrobustcmd{\boldmath}{}
    \caption{
        \label{tab:fvd}
        \textbf{FVD Scores on KITTI and Cityscapes.} }
    \centering
    \small
    \vspace{-2mm}
    \begin{tabular}{lcc}
        \toprule
        Dataset & KITTI & Cityscapes \tabularnewline
        \midrule
        SVG~\cite{Denton2018ICML} & 1733 $\pm$ 198 & 870 $\pm$ 95 \tabularnewline
        SRVP~\cite{Franceschi2020ICML} & 1792 $\pm$ 190 &  1409 $\pm$ 138 \tabularnewline
        SLAMP~\cite{Akan2021ICCV} & 1585 $\pm$ 154 & 796 $\pm$ 89 \tabularnewline
        VRNN~\cite{Castrejon2019ICCV} & \bfseries 1022 $\pm$ 145 & \bfseries 658 $\pm$ 80 \tabularnewline
        Ours-Comb. & 1463 $\pm$ 186 & 793 $\pm$ 86 \tabularnewline
        Ours-Cond. & \underline{1297 $\pm$ 142} & \underline{789 $\pm$ 84} \tabularnewline
    \end{tabular}
    \vspace{-3mm}
\end{table}

\boldparagraph{Video-Level Evaluations} We also use a video level evaluation metric, FVD~\cite{Unterthiner2019ARXIV}, for a video-level comparison. According to the results in \tabref{tab:fvd}, VRNN performs the best in terms of FVD. Our models, both combined and conditional, outperform all the other methods and approach the performance of VRNN with a nearly $40$ times shorter inference time. 

\begin{figure*}[ht!]
\centering
\includegraphics[width=\textwidth]{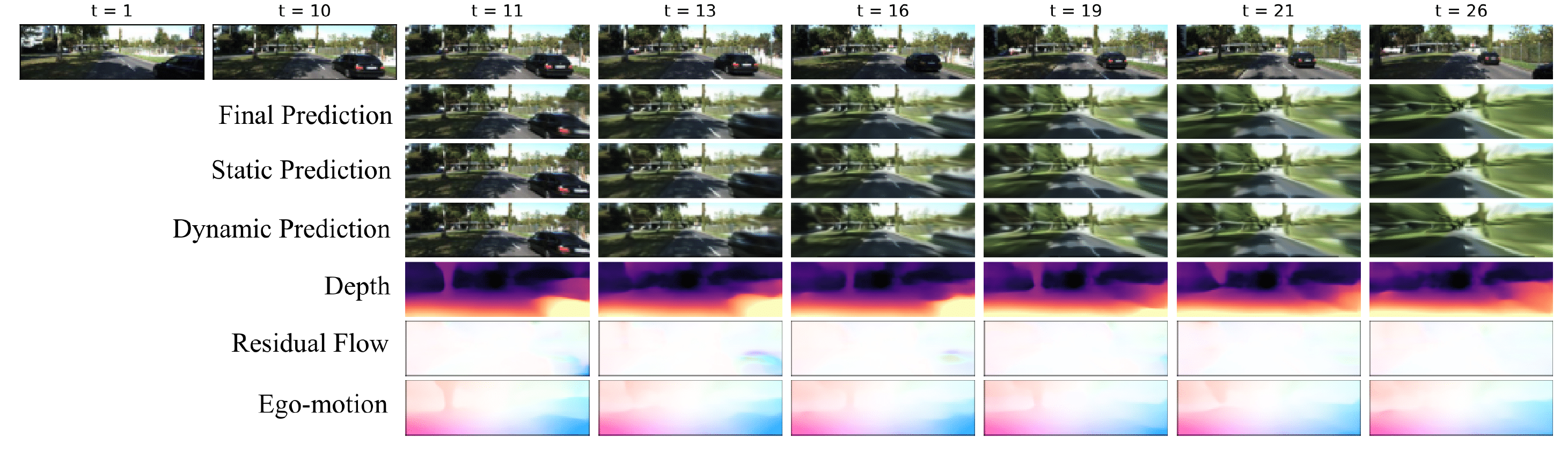}
\caption{\textbf{Conditional Model on KITTI.}
The top row shows the ground truth, followed by the frame predictions by the final, the static, and the dynamic. The last three rows show depth, residual flow, and flow due to ego-motion.
The structure and camera motion can explain the static part of the scene and the independent motion of the moving car on the left is captured by the residual flow.
}
\label{fig:qual_detailed}
\end{figure*}

\begin{table*}[h!]
    \sisetup{detect-weight, table-align-uncertainty=true, mode=text}
    \renewrobustcmd{\bfseries}{\fontseries{b}\selectfont}
    \renewrobustcmd{\boldmath}{}
    \centering
    \caption{\label{tab:fg_bg_unified}
        \textbf{Foreground and Background Evaluations on Cityscapes and KITTI.} We use off-the-shelf semantic segmentation network to identify the foreground regions, mask all the other regions and compute the metrics separately in the foreground and the background.}
    \label{tab:kitti_city_results}
\begin{subtable}[h]{\linewidth}
    \resizebox{\textwidth}{!}{\begin{tabular}{lccc|ccc}%
        \toprule
        \multirow{2}{*}{Models} &  \multicolumn{3}{c}{Foreground} & \multicolumn{3}{c}{Background} \\ \cmidrule{2-7}
        {} & PSNR ($\uparrow$) & SSIM ($\uparrow$) & LPIPS ($\downarrow$) & PSNR ($\uparrow$) & SSIM ($\uparrow$) & LPIPS ($\downarrow$) \tabularnewline 
        \midrule
        SVG \cite{Denton2018ICML} & 30.63 $\pm$ 0.14 & 0.9483 $\pm$ 0.0006 &  0.0676 $\pm$ 0.0005 & 21.24 $\pm$ 0.04 & 0.6699 $\pm$ 0.0014 & 0.2620 $\pm$ 0.0011 \tabularnewline
        SRVP \cite{Franceschi2020ICML}  & 30.85 $\pm$ 0.14 & 0.9474 $\pm$ 0.0006 & 0.0763 $\pm$ 0.0006 & 21.96 $\pm$ 0.04 & 0.6757 $\pm$ 0.0014 & 0.3358 $\pm$ 0.0012 \tabularnewline
        SLAMP~\cite{Akan2021ICCV} &  \underline{31.71} $\pm$ 0.15 &  \underline{0.9536} $\pm$ 0.0005 &  0.0577 $\pm$ 0.0005 & \bfseries 22.66 $\pm$ 0.05 & \bfseries 0.7087 $\pm$ 0.0014 & \underline{0.2341} $\pm$ 0.0011 \tabularnewline
        Improved-VRNN \cite{Castrejon2019ICCV} & 30.65 $\pm$ 0.14 & 0.9495 $\pm$ 0.0006 & \bfseries 0.0554 $\pm$ 0.0004 &  \underline{22.52} $\pm$ 0.07 &  0.6811 $\pm$ 0.0017 &  \bfseries 0.2155 $\pm$ 0.0013 \tabularnewline
        Ours-Combined & 31.52 $\pm$ 0.14 & \underline{0.9536} $\pm$ 0.0005 &  \underline{0.0568} $\pm$ 0.0004 & 21.87 $\pm$ 0.04 &  0.6935 $\pm$ 0.0013 & 0.2496 $\pm$ 0.0009 \tabularnewline
        Ours-Conditional & \bfseries 31.77 $\pm$ 0.14 &  \bfseries 0.9542 $\pm$ 0.0005 & 0.0579 $\pm$ 0.0005 &  22.36 $\pm$ 0.05 & \underline{0.7026} $\pm$ 0.0013 &  0.2426 $\pm$ 0.0009 \tabularnewline
        \bottomrule
    \end{tabular}}
    \vspace{1mm}
    \subcaption{Cityscapes~\cite{Cordts2016CVPR}}
    \label{tab:fg_bg_results_city}
    \vspace{2mm}
\end{subtable}
\begin{subtable}[h]{\linewidth}
    \resizebox{\textwidth}{!}{\begin{tabular}{lccc|ccc}
        \toprule
        \multirow{2}{*}{Models} &  \multicolumn{3}{c}{Foreground} & \multicolumn{3}{c}{Background} \\ \cmidrule{2-7}
        {} & PSNR ($\uparrow$) & SSIM ($\uparrow$) & LPIPS ($\downarrow$) & PSNR ($\uparrow$) & SSIM ($\uparrow$) & LPIPS ($\downarrow$) \tabularnewline 
        \midrule
        SVG \cite{Denton2018ICML} & 26.47 $\pm$ 0.31 & 0.9147 $\pm$ 0.0026 &  0.1183 $\pm$ 0.0027 & 13.30 $\pm$ 0.07 & 0.4003 $\pm$ 0.0034 & 0.5227 $\pm$ 0.0031 \tabularnewline
        SRVP \cite{Franceschi2020ICML} & \underline{27.28} $\pm$ 0.31 & 0.9162 $\pm$ 0.0026 & 0.1244 $\pm$ 0.0029 & 13.97 $\pm$ 0.06 & 0.4080 $\pm$ 0.0033 & 0.5498 $\pm$ 0.0028 \tabularnewline
        SLAMP~\cite{Akan2021ICCV} &  26.93 $\pm$ 0.32 &  0.9154 $\pm$ 0.0026 &  0.1125 $\pm$ 0.0026 & 14.09 $\pm$ 0.08 & 0.4089 $\pm$ 0.0032 & 0.4707 $\pm$ 0.0029 \tabularnewline
        Improved-VRNN \cite{Castrejon2019ICCV} & 26.86 $\pm$ 0.27 & \underline{0.9192} $\pm$ 0.0025 & \bfseries 0.0877 $\pm$ 0.0020 &  14.82 $\pm$ 0.07 &  0.4417 $\pm$ 0.0037 &  \bfseries 0.3380 $\pm$ 0.0031 \tabularnewline
        Ours-Combined & \bfseries 27.44 $\pm$ 0.29 &  \bfseries 0.9193 $\pm$ 0.0025 &  0.1093 $\pm$ 0.0024 & \bfseries 15.11 $\pm$ 0.06 & \underline{0.4455} $\pm$ 0.0032 & 0.4712 $\pm$ 0.0028 \tabularnewline
        Ours-Conditional & 26.80 $\pm$ 0.29 & 0.9161 $\pm$ 0.0026 & \underline{0.1046} $\pm$ 0.0023 &  \underline{15.02} $\pm$ 0.05 & \bfseries 0.4507 $\pm$ 0.0031 &  \underline{0.4451} $\pm$ 0.0028 \tabularnewline
        \bottomrule
    \end{tabular}}
    \vspace{1mm}
    \subcaption{KITTI~\cite{Geiger2012CVPR, Geiger2013IJRR}}
    \label{tab:fg_bg_results_kitti}
\end{subtable}
\vspace{-0.5cm}
\end{table*}

\boldparagraph{Foreground and Background Evaluations}
We compare the prediction performance separately in the foreground and the background regions of the scene in \tabref{tab:kitti_city_results} on KITTI (\subref{tab:fg_bg_results_kitti}) and Cityscapes (\subref{tab:fg_bg_results_city}). 
We use off-the-shelf semantic segmentation models to extract the objects in the scene and assume some of the semantic classes as the foreground. Specifically, we use the pre-trained model by \cite{Zhu2019CVPR, Reda2018ECCV}, which is the best model on KITTI semantic segmentation leaderboard with code available (the second-best overall), to obtain the masks on this dataset. On Cityscapes, we use the pre-trained model by \cite{Tao2020Arxiv} for similar reasons. We choose the following classes as the foreground objects: ``person, rider, car, truck, bus, train, motorcycle, bicycle" and consider the remaining pixels as the background.
For foreground results, we extract the foreground regions based on the segmentation result and ignore all the other pixels in the background by assigning the mean color, gray. 
We do the opposite masking for the background region by setting all of the foreground regions to gray, and calculate the metrics again. 

On Cityscapes, our \textit{Combined} and \textit{Conditional} models achieve the best results in terms of PSNR and SSIM metrics in the foreground regions. The \textit{Conditional} outperforms the \textit{Combined}, showing the importance of conditioning of dynamic latent variables on the static for modelling the independent motion of foreground objects. However, SLAMP and Improved-VRNN outperform our models in the background. Improved-VRNN's performance is especially impressive in terms of LPIPS, consistently in all regions on both datasets.
On KITTI, our \textit{Combined} model performs the best in foreground regions in terms of PSNR and SSIM. For background regions, our \textit{Combined} and \textit{Conditional} models are the two best-performing models in terms of PSNR and SSIM.
In summary, we validate our two claims; first, to better model the motion history of foreground objects on the more dynamic Cityscapes and second, to better model the larger motion in the background on KITTI due to a smaller frame rate.

\begin{figure*}[h!]
\centering
\includegraphics[width=\textwidth]{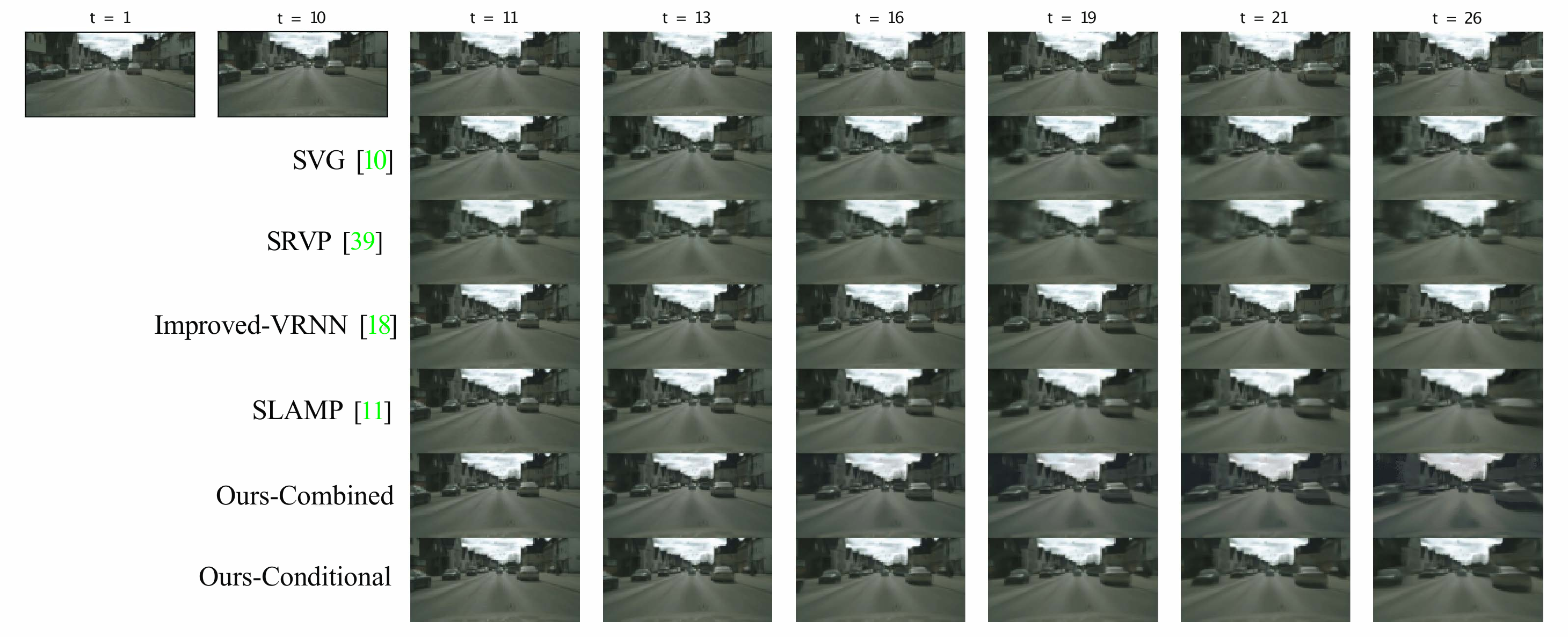}
\caption{\textbf{Comparison to Other Methods on Cityscapes.} 
Our \emph{Conditional} model can better predict the motion of the vehicles moving on the left and on the right thanks to the separate modeling of the motion history of background and foreground regions.
}
\label{fig:qual_comp}
\vspace{-0.5cm}
\end{figure*}

\begin{figure*}[h!]
\centering
\includegraphics[width=\textwidth]{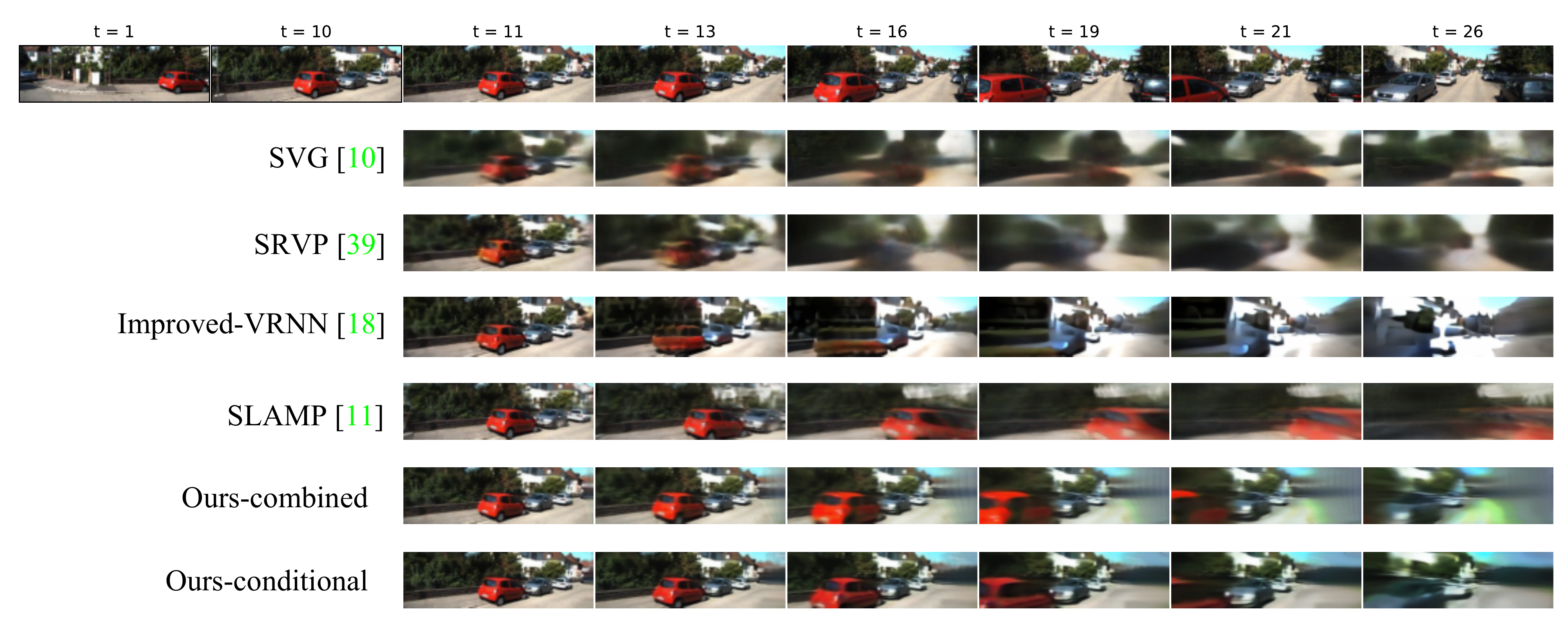}
\caption{\textbf{Difficult Turning Case on KITTI.} 
Our \emph{Conditional} and \emph{Combined} model can better predict future after the conditioning frames. Our models, especially the conditional model, can better preserve the red car in their predictions for a longer time. All of the other methods fail in the later time steps due to the difficulty of predicting future in the turning case which is not frequent in the dataset.
}
\label{fig:qual_comp_kitti}
\end{figure*}

\boldparagraph{Evaluation of Depth Predictions}
\begin{table*}[h!]
    \caption{
        \label{tab:depth_eval}
        \textbf{Evaluation of Depth Predictions.} We compare depth predictions from our \emph{Combined} and \emph{Conditional} models to the state-of-the-art unsupervised monocular depth prediction method Monodepth2~\cite{Godard2019ICCV} on KITTI. We predict future depth without seeing the target frame.
    }
    \sisetup{detect-weight, table-align-uncertainty=true, mode=text}
    \renewrobustcmd{\bfseries}{\fontseries{b}\selectfont}
    \renewrobustcmd{\boldmath}{}
    \centering
    \vspace{-0.1in}
    \resizebox{\textwidth}{!}{\begin{tabular}{l|ccccccc}%
        \toprule
        Models &  Abs Rel ($\downarrow$) & Sq Rel ($\downarrow$) & RMSE ($\downarrow$) & RMSE$_{log}$ ($\downarrow$) & $\delta < 1.25$ ($\uparrow$) & $\delta < 1.25^2$ ($\uparrow$) & $\delta < 1.25^3$ ($\uparrow$) \tabularnewline 
        \midrule
        Ours-Combined & 0.204 & 2.388 & 7.232 & 0.289 & 0.729 & 0.892 & 0.949
        \tabularnewline
        Ours-Conditional & 0.221 & 3.173 & 7.474 & 0.298 & 0.724 & 0.887 & 0.944
        
        \tabularnewline
        Monodepth2~\cite{Godard2019ICCV} & \bfseries 0.136 & \bfseries 1.095 & \bfseries 5.369 & \bfseries 0.213 & \bfseries 0.836 & \bfseries 0.946 & \bfseries 0.977 
        \tabularnewline
        \bottomrule
    \end{tabular}}
    \vspace{-0.35cm}
\end{table*}
In order to validate the quality of our depth predictions, we evaluate them on KITTI with respect to a state-of-the-art monocular depth estimation approach Monodepth2~\cite{Godard2019ICCV} by training it on a similar resolution to ours ($320 \times 96$). Monodepth2 serves as an upper-bound for the performance of our models because we predict future depth based on previous frame predictions without actually seeing the target frame. %
As can be seen from \tabref{tab:depth_eval}, both our models perform reasonably well in all metrics~\cite{Eigen2014NeurIPS} in comparison to Monodepth2.

\begin{figure*}[ht!]
\centering
\includegraphics[width=\textwidth]{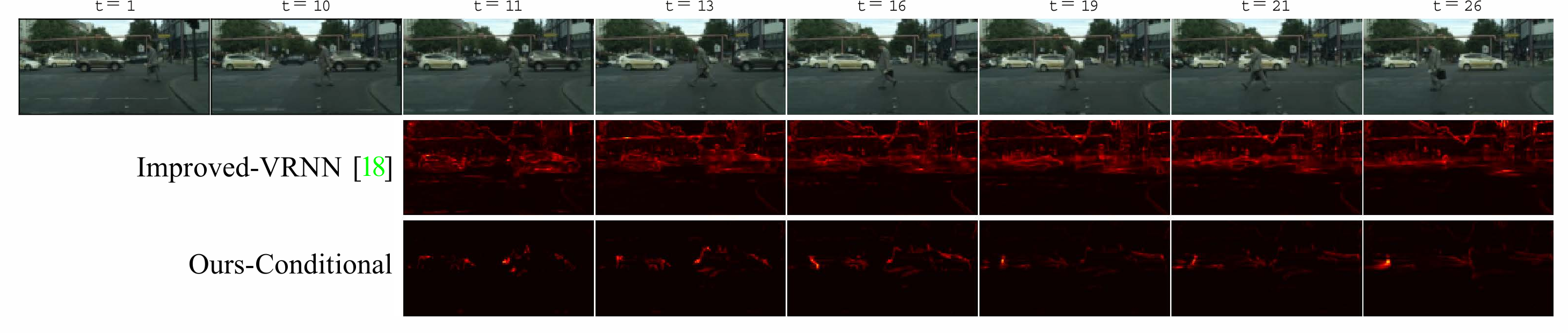}
\caption{\textbf{Diversity of Samples on Cityscapes.} We compare the standard deviations of 100 samples generated by Improved-VRNN~\cite{Castrejon2019ICCV} and our \emph{Conditional} model. Our model can pinpoint uncertain regions around the foreground objects in comparison to more uniform results of Improved-VRNN.}
\label{fig:diversity}
\vspace{-0.5cm}
\end{figure*}

\begin{table}[t!]
    \caption{\textbf{Run-time Comparison of the Models.} We measure the time needed to generate 10 samples and report the results in seconds as the average over the dataset. We only measure the GPU processing time and discard the other operations.}
    \label{tab:speed}
    \sisetup{detect-weight, table-align-uncertainty=true, mode=text}
    \centering
    \vspace{0.1in}
    \begin{tabular}{lrr}%
        \toprule
        \textbf{Models} &  \textbf{KITTI} & \textbf{Cityscapes} \tabularnewline 
        \midrule
        SVG~\cite{Denton2018ICML} & 1.04 & 0.90
        \tabularnewline
        SRVP~\cite{Franceschi2020ICML}  & 0.11 & 0.12 \tabularnewline
        Improved VRNN~\cite{Castrejon2019ICCV}  & 39.25 & 41.59 \tabularnewline
        SLAMP~\cite{Akan2021ICCV}  & 2.03 & 1.44 \tabularnewline
        Ours-Combined & 1.02 & 1.16 \tabularnewline
        Ours-Conditional & 1.03 & 1.21 \tabularnewline
        \bottomrule
    \end{tabular}
    \vspace{-5mm}
\end{table}
\boldparagraph{Run-time comparisons}
We compare the methods in terms of their run-time in \tabref{tab:speed}.
We measure the time needed to generate 10 samples and report the results in seconds as the average over the dataset. There is a trade off between the run-time and the prediction performance of the models. Improved-VRNN has the largest run-time by far due to its hierarchical latent space, with nearly 40 seconds of processing time for ten samples. The performance of Improved-VRNN is impressive, especially in terms of both LPIPS and FVD, however costly in terms of both memory and run-time, therefore not applicable to autonomous driving. Our models, both Combined and Conditional, have a similar run-time on KITTI and only a slight increase in run-time compared to vanilla stochastic video generation model SVG~\cite{Denton2018ICML}. The state-space model SRVP~\cite{Franceschi2020ICML} leads to the best run-time by removing the need for autoregressive predictions but at the cost of low performance on real-world datasets.

\subsection{Qualitative Results}
We visualize the results of our model including the intermediate predictions on KITTI (\figref{fig:qual_detailed}) as well as the final prediction in comparison to previous methods on Cityscapes (\figref{fig:qual_comp}). Please see Appendix for more results on both datasets. 
As can be seen from \figref{fig:qual_detailed}, our model can predict structure and differentiate between the two types of motion in the scene. First, the camera motion in the background is predicted~(\textit{Ego-motion}) and the motion of the foreground object is predicted as residual on top of it~(\textit{Residual Flow}).
The scene decomposition into static and dynamic and separate modeling of motion in each allow our model to generate better predictions of future in dynamic scenes. \figref{fig:qual_comp} shows a case where our model can correctly estimate the change in the shape and the size of the two vehicles moving in comparison to previous approaches.

We evaluate the diversity of predictions qualitatively in \figref{fig:diversity}. For this purpose, we visualize the standard deviations of generated frames over 100 samples for Improved-VRNN~\cite{Castrejon2019ICCV} and our \emph{Conditional} model. The higher the standard deviation at a pixel, the higher the uncertainty is at that pixel due to differences in samples. While the uncertainty is mostly uniform over the image for Improved-VRNN, our model can pinpoint it to the foreground regions thanks to our separate modeling of the motion history in the background and the foreground regions.

\section{Conclusion and Future Work}
\label{sec:conc}
We introduced a conditional stochastic video prediction model by decomposing the scene into static and dynamic in driving videos. We showed that separate modelling of foreground and background motion leads to better future predictions. Our method is among the top-performing methods overall while still being efficient. The modelling of the static part with domain knowledge is justified on KITTI with large camera motion. The separate modelling of foreground objects as residual motion leads to better results in dynamic scenes of Cityscapes. Our visualizations show that flow prediction can capture the residual motion of foreground objects on top of ego-motion thanks to our conditional model.

We found that modelling the stochasticity in terms of underlying factors is beneficial for autonomous driving. From this point forward, we see three important directions to improve stochastic video prediction in autonomous driving. Since the uncertainty is mostly due to foreground objects, stochastic methods can focus on foreground objects with more sophisticated models. Next, there is a limitation in our model due to the warping operation. Our model can predict the future motion of a car visible in the conditioning frames but it cannot predict a car appearing after that. For that, we need other ways of obtaining self-supervision without warping. Finally, we could not use any of the well-known tricks in view synthesis such as multi-scale, forward-backward consistency check, 3D convolutions, and better loss functions~\cite{Godard2019ICCV, Guizilini2020CVPR} due to computational reasons. Any improvement there can yield performance gains for stochastic video prediction with structure and motion.
\vspace{-0.3cm}


\bibliographystyle{IEEEtran}
\bibliography{bibliography_long, ref}

\ifCLASSOPTIONcompsoc
  \section*{Acknowledgments}
\else
  \section*{Acknowledgment}
\fi

Adil Kaan Akan and Sadra Safadoust were supported by KUIS AI Center fellowship, Fatma G\"{u}ney by ``TUBITAK 2232 International Fellowship for Outstanding Researchers Programme" by TÜBİTAK and the ``Marie Skłodowska-Curie Individual Fellowship" by the European Commission. \vspace{-0.5cm}
\begin{IEEEbiography}[{\includegraphics[trim={0 2cm 0 0},clip, width=1.0\textwidth]{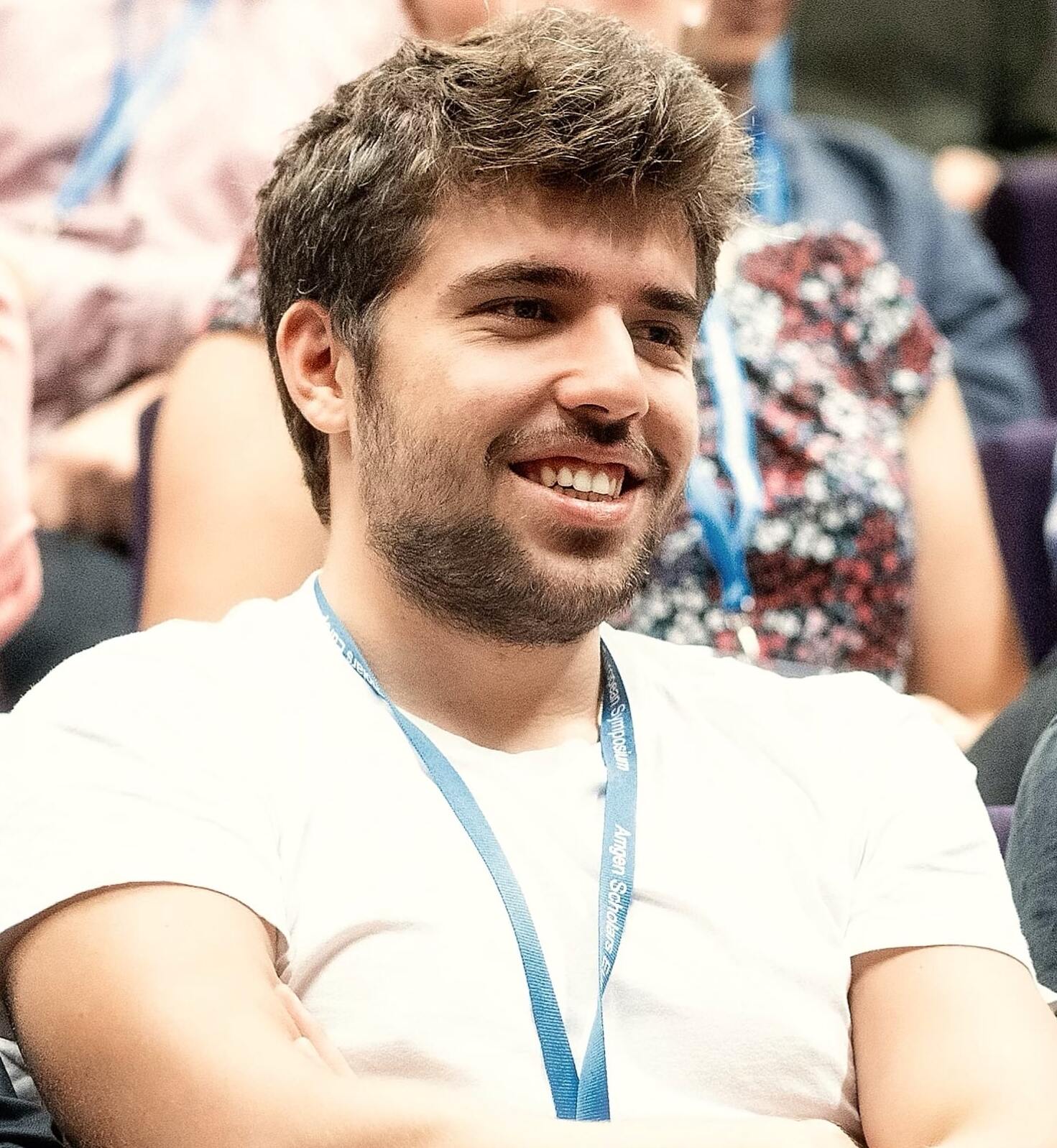}}]{Adil Kaan Akan}
is an M.Sc. student in Koç University. He received his B.Sc. degree in computer science from Middle East Technical University, Turkey. His research interests include machine learning and computer vision with special interest in stochastic future prediction and video understanding. \vspace{-0.3cm}
\end{IEEEbiography}
\begin{IEEEbiography}[{\includegraphics[width=1.0\textwidth]{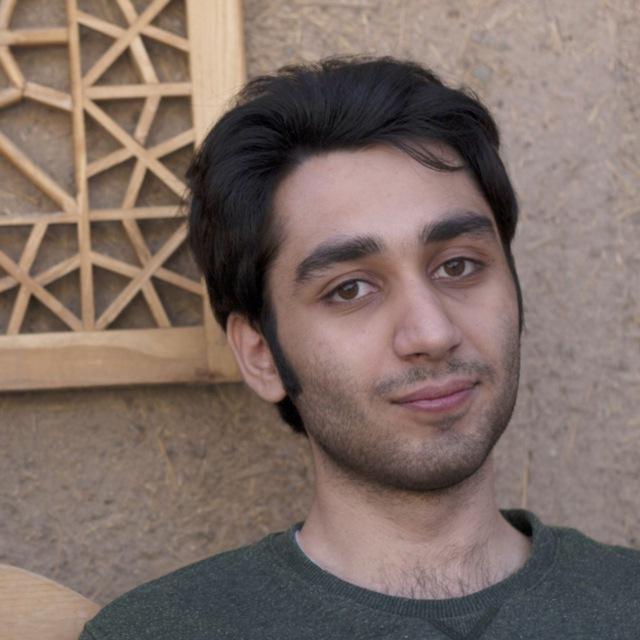}}]{Sadra Safadoust}
 He received the B.Sc. degree in computer engineering from Sharif University of Technology, Iran. He is an M.Sc. student at Koç University, Turkey. His research interests include computer vision and deep learning, with a special interest in depth estimation and stochastic future prediction. \vspace{-0.3cm}
\end{IEEEbiography}
\begin{IEEEbiography}[{\includegraphics[width=1.0\textwidth]{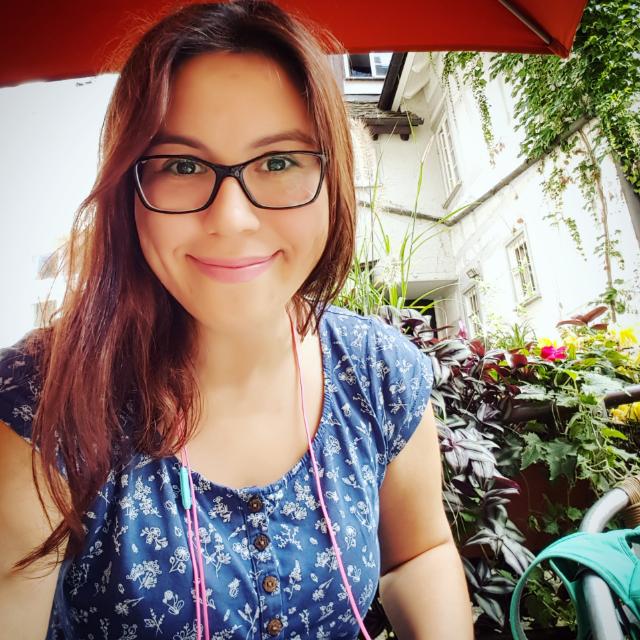}}]{Fatma Güney}
 is an Assistant Professor at the Department of Computer Engineering and a researcher at the KUIS AI center in Istanbul. Before joining KUIS AI, she received her Ph.D. from the Max Planck Institute for Intelligent Systems and worked as a postdoctoral researcher at the University of Oxford. In the last couple of years, she has been awarded the International Fellowship for Outstanding Researchers by TÜBİTAK, the Marie Skłodowska-Curie Individual Fellowship by the European Commission, and the Newton Advanced Fellowship by the British Royal Society. She is a recipient of multiple outstanding reviewer awards at the leading computer vision conferences. Her research interests include 3D computer vision and representation learning from video sequences.
\end{IEEEbiography}

\onecolumn
\newpage
\begin{appendices}
In this part, we provide derivations, model details, and training settings, and show additional results for our paper ``Stochastic Video Prediction with Structure and Motion".
We first provide the full derivation of evidence lower bound of our model in \secref{sec:derivations}. 
We explain the details of different architectures, training details for our models, and changes applied to the baselines in \secref{sec:training_details}.
In \secref{sec:results}, we include across time quantitative results and additional qualitative comparisons to other methods, detailed visualizations of our model's components, and additional diversity examples.

\section{Derivations}
\label{sec:derivations}
In this section, we derive the variational lower bound for the proposed method with the following generative model:
\begin{equation}
\label{eq:generative_model}
p_{\btheta}(\bx) = \iint p(\bx | \bz^s, \bz^d ) p(\bz^s | \bx) p(\bz^d | \bx) \mathrm{d}\bz^s \mathrm{d}\bz^d.
\end{equation}

We assume that the inference of the dynamic part $\bz^d$ depends on the static part $\bz^s$:
\begin{equation}
\label{eq:posterior_prior}
p(\bz^s | \bx) \approx q_s(\bz^s | \bx)~~\text{and}~~\: p(\bz^d | \bx) \approx q_d(\bz^d | \bx, \bz^s) .
\end{equation}

The log-posterior then becomes:
\begin{align}
\label{eq:without_time}
\log p_{\theta}(\bx) &= \log \left( \iint p(\bx | \bz^s, \bz^d) p(\bz^s | \bx) p(\bz^d | \bx) \mathrm{d}\bz^s \mathrm{d}\bz^d \right) \nonumber \\
&= \log \left( \iint p(\bx | \bz^s, \bz^d) p(\bz^s | \bx) p(\bz^d | \bx) \dfrac{q_d(\bz^d | \bz^s, \bx)}{q_d(\bz^d | \bz^s, \bx)} \dfrac{q_s(\bz^s | \bx)}{q_s(\bz^s | \bx)} \mathrm{d}\bz^s \mathrm{d}\bz^d \right) \nonumber \\
&= \log \left( \int \mathbb{E}_{q_d(\bz^d | \bz^s, \bx)}\left[ \dfrac{p(\bx | \bz^s, \bz^d) p(\bz^s | \bx)}{q_d(\bz^d | \bz^s, \bx)} \right] p(\bz^s | \bx) \dfrac{q_s(\bz^s | \bx)}{q_s(\bz^s | \bx)} \mathrm{d}\bz^s \right) \nonumber \\
&=\log \mathbb{E}_{q_s(\bz^s | \bx)} \left[ \mathbb{E}_{q_d(\bz^d | \bz^s, \bx)}\left[ \dfrac{p(\bx | \bz^s, \bz^d) p(\bz^d | \bx)}{q_d(\bz^d | \bz^s, \bx)} \right] \dfrac{p(\bz^s | \bx)}{q_s(\bz^s |\bx)} \right]
\end{align}

By using Jensen's inequality once to exchange the outer expectation with the $\log$ leads to the following equation:
\begin{align}
\log p_{\btheta}(\bx) &\geq \mathbb{E}_{q_s(\bz^s | \bx)} \left[ \log \left( \mathbb{E}_{q_d(\bz^d | \bz^s, \bx)}\left[ \dfrac{p(\bx | \bz^s, \bz^d) p(\bz^d | \bx)}{q_d(\bz^d | \bz^s, \bx)} \right] \right) + \log \left(\dfrac{p(\bz^s | \bx)}{q_s(\bz^s |\bx)} \right) \right] \nonumber \\
&=\mathbb{E}_{q_s(\bz^s | \bx)} \left[ \log \left( \mathbb{E}_{q_d(\bz^s | \bz^s, \bx)}\left[ \dfrac{p(\bx | \bz^s, \bz^d) p(\bz^d | \bx)}{q_d(\bz^d | \bz^s, \bx)} \right] \right) \right] - D_{KL}(q_s(\bz^s | \bx) || p(\bz^s | \bx)) 
\end{align}

Then, by using Jensen's inequality once more to exchange the $\log$ and the inner expectation, we obtain the following:
{\small
\begin{align}
\log p_{\btheta}(\bx) \geq& \mathbb{E}_{q_s(\bz^s | \bx)} \left[ \mathbb{E}_{q_d(\bz^d | \bz^s, \bx)}\left[ \log p(\bx | \bz^s, \bz^d) + \log \left(\dfrac{p(\bz^d | \bx)}{q_d(\bz^d | \bz^s, \bx)} \right] \right) \right] - D_{KL}(q_s(\bz^s | \bx) || p(\bz^s | \bx)) \nonumber \\
=&\underbrace{\mathbb{E}_{q_s(\bz^s | \bx)} \left[ \mathbb{E}_{q_d(\bz^d | \bz^s, \bx)}\left[ \log p(\bx | \bz^s, \bz^d)\right] \right]}_\text{1} - \underbrace{\mathbb{E}_{q_s(\bz^s | \bx)} \left[D_{KL}(q_d(\bz^d | \bz^s, \bx) || p(\bz^d | \bx)) \right]}_\text{2} - \nonumber \\ &\underbrace{D_{KL}(q_s(\bz^s | \bx) || p(\bz^s | \bx))}_\text{3} 
\end{align}
}
where the first term is the reconstruction error between $\bx$ and $p(\bx | \bz^d , \bz^s)$. The second term is the KL divergence for the dynamic part $q_d(\bz^d | \bz^s, \bx)$ with the prior $p(\bz^d | \bx)$ and the last term is the KL divergence for the static part $q_s(\bz^s | \bx)$ with the prior $p(\bz^s | \bx)$. %

We model the posterior distributions with two recurrent networks. The recurrent networks output two different posterior distributions, $q_{s}(\bz^{s}_t \vert \bx_{1:t})$ and $q_{d}(\bz^{d}_t \vert \bz^s_{1:t}, \bx_{1:t})$, at every time step. Due to the independence assumption of the latent variables across time, $\bz^{s} = [\bz^{s}_1, \bz^{s}_2, \cdots, \bz^{s}_T]$ and $\bz^{d} = [\bz^{d}_1, \bz^{d}_2, \cdots, \bz^{d}_T]$, we can derive the estimation of posterior distributions across time steps as follows:
\begin{align}
    q_{s}(\bz^s \vert \bx) &= \prod\limits_t q_{s}(\bz^{s}_t \vert \bx_{1:t})  \nonumber \\
    q_{d}(\bz^d \vert \bx) &= \prod\limits_t q_{d}(\bz^{d}_t \vert \bz^s_{1:t}, \bx_{1:t})
\end{align}

Since the latent variables, $\bz^{s} = [\bz^{s}_1, \bz^{s}_2, \dots, \bz^{s}_T]$ and $\bz^{d} = [\bz^{d}_1, \bz^{d}_2, \dots, \bz^{d}_T]$, are independent across time and independent from each other, we can further decompose Kullback-Leibler terms in the evidence lower bound into individual time steps.

At each time step, our model predicts $\bx_{t}$, conditioned on $\bx_{t-1}$, $\bz^{s}_{t}$, and $\bz^{d}_{t}$. Since our model has recurrence connections, it considers not only $\bx_{t-1}$,  $\bz^{s}_{t}$ and $\bz^{d}_{t}$, but also $\bx_{1:t-2}$, $\bz^{s}_{1:t-1}$, and $\bz^{d}_{1:t-1}$. Therefore, we can further write our inference as follows:
\begin{align}
    \log p_{\btheta}(\bx \vert \bz^{s}, \bz^{d}) &= \log \prod\limits_t p_{\btheta}(\bx_t \vert \bx_{1:t-1}, \bz^{s}_{1:t}, \bz^{d}_{1:t})\\
    &= \sum\limits_t \log p_{\btheta}(\bx_t \vert \bx_{1:t-1}, \bz^{s}_{1:t}, \bz^{d}_{1:t}) \nonumber
\end{align}

Combining all of them leads to the following variational lower bound:
{\small
\begin{align}
\label{eq:final_elbo_conditional}
\log p_{\btheta}(\bx) &\geq \mathcal{L}(\bx_{1:T}) \\
&=  \mathbb{E}_{\substack{\bz^s \sim q_{s} \\
                                   \bz^d \sim q_{d}}} \log p_{\btheta}(\bx \vert \bz^{s}, \bz^{d}) - D_{\mathrm{KL}}q_{s}(\bz^{s} \vert \bx) \mid \mid p_{s}(\bz^{s} \vert \bx)) - D_{\mathrm{KL}}(q_{d}(\bz^{d} \vert \bz^s, \bx) \mid \mid p_{d}(\bz^{d} \vert \bz^s, \bx)) \nonumber \\
&=  \sum\limits_t \mathbb{E}_{\substack{\bz^s \sim q_{s} \\
                                   \bz^d \sim q_{d}}}   \log p_{\btheta}(\bx_t \vert \bx_{1:t-1}, \bz^{s}_{1:t}, \bz^{d}_{1:t}) \nonumber \\
& \hspace{15mm} -  D_{\mathrm{KL}}(q_{s}(\bz^{s}_t \vert \bx_{1:t}) \mid \mid p_{s}(\bz^{s}_t \vert \bx_{1:t-1})) \nonumber \\
& \hspace{15mm} - D_{\mathrm{KL}}(q_{d}(\bz^{d}_t \vert \bz^{s}_{1:t},\bx_{1:t}) \mid \mid p_{d}(\bz^{d}_t \vert \bz^s_{1:t-1}, \bx_{1:t-1})) \nonumber
\end{align}
}

The \textit{Combined} model splits the latent space into two without relating them to each other.
We can obtain the ELBO of the \textit{Combined} model by removing the dependency of the static on the dynamic: $q_d(\bz^d | \bz^s, \bx) = q_d(\bz^d | \bx)$. This leads to independent latent variables with the following ELBO: 

{\small
\begin{align}
\label{eq:final_elbo_conditional2}
\log p_{\btheta}(\bx) & \geq \mathcal{L}(\bx_{1:T}) \\
&=  \mathbb{E}_{\substack{\bz^s \sim q_{s} \\
                                  \bz^d \sim q_{d}}} \log p_{\btheta}(\bx \vert \bz^{s}, \bz^{d}) - D_{\mathrm{KL}}(q_{s}(\bz^{s} \vert \bx) \mid \mid p_{s}(\bz^{s} \vert \bx)) - D_{\mathrm{KL}}(q_{d}(\bz^{d} \vert \bz^s, \bx) \mid \mid p_{d}(\bz^{d} \vert \bz^s, \bx)) \nonumber \\
&=  \sum\limits_t \mathbb{E}_{\substack{\bz^s \sim q_{s} \\
                                  \bz^d \sim q_{d}}}   \log p_{\btheta}(\bx_t \vert \bx_{1:t-1}, \bz^{s}_{1:t}, \bz^{d}_{1:t}) \nonumber \\
& \hspace{15mm} -  D_{\mathrm{KL}}(q_{s}(\bz^{s}_t \vert \bx_{1:t}) \mid \mid p_{s}(\bz^{s}_t \vert \bx_{1:t-1})) \nonumber \\
& \hspace{15mm} - D_{\mathrm{KL}}(q_{d}(\bz^{d}_t \vert \bx_{1:t}) \mid \mid p_{d}(\bz^{d}_t \vert \bx_{1:t-1})) \nonumber
\end{align}
}

\section{Model Details and Training Settings}
\label{sec:training_details}
\subsection{Our Models}
\begin{table}[!b]
\caption{\textbf{VGG Encoder Architecture.} This table shows the details of the encoder architecture for the VGG model. The columns $\bk,\bs, \bp$ stand for kernel size, stride, and padding, respectively. The resolution shows the feature map resolution for KITTI and Cityscapes separately.}
\vspace{0.1in}
\label{tab:enc}
\centering
\begin{tabular}{|c|c|c|c|c|c|c}
\hline
\multicolumn{7}{|c|}{\textbf{VGG-based Encoder} }                                                                              \\ \hline
\textbf{Blocks} & \textbf{k} & \textbf{s} & \textbf{p} & \textbf{Channels} & \textbf{Resolution} & \multicolumn{1}{c|}{\textbf{Act.}} \\ \hline
Conv         & 3          & 2          & 1          & 64           & $46 \times 156$  | $64 \times 128$     & \multicolumn{1}{c|}{Leaky ReLU} \\
Conv         & 3          & 1          & 1          & 64           & $46 \times 156$     | $64 \times 128$      & \multicolumn{1}{c|}{Leaky ReLU} \\ \hline
Conv         & 3          & 2          & 1          & 96           & $24 \times 80$   | $32 \times 64$        & \multicolumn{1}{c|}{Leaky ReLU} \\
Conv         & 3          & 1          & 1          & 96           & $24 \times 80$   | $32 \times 64$         & \multicolumn{1}{c|}{Leaky ReLU} \\ \hline
Conv         & 3          & 2          & 1          & 128           & $12 \times 40$ | $16 \times 32$ & \multicolumn{1}{c|}{Leaky ReLU} \\
Conv         & 3          & 1          & 1          & 128           & $12 \times 40$ | $16 \times 32$      & \multicolumn{1}{c|}{Leaky ReLU} \\ 
Conv         & 3          & 1          & 1          & 128           & $12 \times 40$  | $16 \times 32$      & \multicolumn{1}{c|}{Leaky ReLU} \\ \hline
Conv         & 3          & 2          & 1          & 196           & $6 \times 20$   | $8 \times 16$      & \multicolumn{1}{c|}{Leaky ReLU} \\
Conv         & 3          & 1          & 1          & 196           & $6 \times 20$     | $8 \times 16$     & \multicolumn{1}{c|}{Leaky ReLU} \\ 
Conv         & 3          & 1          & 1          & 196           & $6 \times 20$     | $8 \times 16$      & \multicolumn{1}{c|}{Leaky ReLU} \\ \hline
Conv         & 3          & 2          & 1          & 256           & $3 \times 10$     | $4 \times 8$       & \multicolumn{1}{c|}{Leaky ReLU} \\
Conv         & 3          & 1          & 1          & 256           & $3 \times 10$       | $4 \times 8$     & \multicolumn{1}{c|}{Leaky ReLU} \\ 
Conv         & 3          & 1          & 1          & 256           & $3 \times 10$      | $4 \times 8$       & \multicolumn{1}{c|}{Leaky ReLU} \\ \hline
Conv         & 3          & 2          & 1          & 128           & $3 \times 10$       | $4 \times 8$       & \multicolumn{1}{c|}{Leaky ReLU} \\
Conv         & 3          & 1          & 1          & 128           & $3 \times 10$      | $4 \times 8$      & \multicolumn{1}{c|}{Leaky ReLU} \\  \hline

\end{tabular}
\end{table}
\begin{table}[!]
\caption{\textbf{VGG Decoder Architecture.} This table shows the details of the decoder architecture for the VGG model. The columns $\bk,\bs, \bp$ stand for kernel size, stride, and padding, respectively. The resolution shows the feature map resolution for KITTI and Cityscapes separately.}
\vspace{0.1in}
\label{tab:dec}
\centering
\begin{tabular}{|c|c|c|c|c|c|c}
\hline
\multicolumn{7}{|c|}{\textbf{VGG-based Decoder}}                                                                              \\ \hline
\textbf{Blocks} & \textbf{k} & \textbf{s} & \textbf{p} & \textbf{Channels} & \textbf{Resolution} & \multicolumn{1}{c|}{\textbf{Act.}} \\ \hline
Conv         & 3          & 1          & 1          & 256           & $3 \times 10$  | $4 \times 8$     & \multicolumn{1}{c|}{Leaky ReLU} \\
Conv         & 3          & 1          & 1          & 256           & $3 \times 10$  | $4 \times 8$     & \multicolumn{1}{c|}{Leaky ReLU} \\ \hline

Conv         & 3          & 1          & 1          & $256\times2$           & $3 \times 10$  | $4 \times 8$        & \multicolumn{1}{c|}{Leaky ReLU} \\
Conv         & 3          & 1          & 1          & 256           & $3 \times 10$  | $4 \times 8$          & \multicolumn{1}{c|}{Leaky ReLU} \\
Conv         & 3          & 1          & 1          & 256           & $3 \times 10$  | $4 \times 8$         & \multicolumn{1}{c|}{Leaky ReLU} \\ \hline

Conv         & 3          & 1          & 1          & $192\times2$           & $6 \times 20$   | $8 \times 16$        & \multicolumn{1}{c|}{Leaky ReLU} \\
Conv         & 3          & 1          & 1          & 192           & $6 \times 20$   | $8 \times 16$        & \multicolumn{1}{c|}{Leaky ReLU} \\
Conv         & 3          & 1          & 1          & 192           & $6 \times 20$   | $8 \times 16$        & \multicolumn{1}{c|}{Leaky ReLU} \\ \hline

Conv         & 3          & 1          & 1          & $128\times2$           & $12 \times 40$   | $16 \times 32$        & \multicolumn{1}{c|}{Leaky ReLU} \\
Conv         & 3          & 1          & 1          & 128           & $12 \times 40$   | $16 \times 32$        & \multicolumn{1}{c|}{Leaky ReLU} \\ \hline

Conv         & 3          & 1          & 1          & $96\times2$           & $24 \times 80$   | $32 \times 64$        & \multicolumn{1}{c|}{Leaky ReLU} \\
Conv         & 3          & 1          & 1          & 96           & $24 \times 80$   | $32 \times 64$         & \multicolumn{1}{c|}{Leaky ReLU} \\ \hline

Conv         & 3          & 1          & 1          & $64\times2$           & $46 \times 156$   | $64 \times 128$        & \multicolumn{1}{c|}{Leaky ReLU} \\
Conv         & 3          & 1          & 1          & 64           & $46 \times 156$   | $64 \times 128$         & \multicolumn{1}{c|}{Leaky ReLU} \\ \hline

Conv         & 3          & 1          & 1          & 64           & $92 \times 310$   | $128 \times 256$        & \multicolumn{1}{c|}{Leaky ReLU} \\
Transposed Conv         & 3          & 1          & 1          & 3           & $92 \times 310$   | $128 \times 256$         & \multicolumn{1}{c|}{Sigmoid} \\ \hline
\end{tabular}
\end{table}
\begin{table}
\caption{\textbf{Low-Resolution Encoder Architecture.} This table show the details of the low-resolution encoder architecture. The columns $\bk,\bs, \bp$ stand for kernel size, stride and padding, respectively. The resolution shows the feature map resolution for KITTI and Cityscapes, sperately.}
\vspace{0.1in}
\label{tab:low_res_enc}
\centering
\begin{tabular}{|c|c|c|c|c|c|c}
\hline
\multicolumn{7}{|c|}{\textbf{Low-Resolution Encoder}}                                                                              \\ \hline
\textbf{Blocks} & \textbf{k} & \textbf{s} & \textbf{p} & \textbf{Channels} & \textbf{Resolution} & \multicolumn{1}{c|}{\textbf{Act.}} \\ \hline
Conv         & 3          & 1          & 1          & 256           & $3 \times 10$  | $4 \times 8$     & \multicolumn{1}{c|}{Leaky ReLU} \\
Conv         & 3          & 1          & 1          & 256           & $3 \times 10$  | $4 \times 8$     & \multicolumn{1}{c|}{Leaky ReLU} \\ 
Squeeze \& Excitation Layer \cite{Hu2018CVPR} & \textemdash &\textemdash &\textemdash &\textemdash & \textemdash     & \multicolumn{1}{c|}{\textemdash} \\ \hline
Conv         & 3          & 1          & 1          & 256           & $3 \times 10$  | $4 \times 8$     & \multicolumn{1}{c|}{Leaky ReLU} \\
Conv         & 3          & 1          & 1          & 128           & $3 \times 10$  | $4 \times 8$     & \multicolumn{1}{c|}{Leaky ReLU} \\ 
Squeeze \& Excitation Layer \cite{Hu2018CVPR} & \textemdash &\textemdash &\textemdash &\textemdash & \textemdash     & \multicolumn{1}{c|}{\textemdash} \\ \hline
Conv         & 3          & 1          & 1          & 64           & $3 \times 10$  | $4 \times 8$     & \multicolumn{1}{c|}{Leaky ReLU} \\
Conv         & 3          & 1          & 1          & 128           & $3 \times 10$  | $4 \times 8$     & \multicolumn{1}{c|}{Leaky ReLU} \\ \hline
\end{tabular}
\end{table}
\boldparagraph{VGG Architecture} For the VGG-based models~\cite{Simonyan2015ICLR}, we respectively use the architectures shown in \tabref{tab:enc} and \tabref{tab:dec} as the encoder and the depth and flow decoders. Here we adopt the pose decoder of \citet{Godard2019ICCV}. We use three low-resolution architectures to encode features for flow, depth, and pose as shown in \tabref{tab:low_res_enc}. The decoder uses skip connections from encoder's feature maps. We simply concatenate intermediate feature maps to the previous layers outputs. For the depth decoder, only one encoded image features are fed to the low resolution encoder; however, for pose and flow features, we feed two consecutive image features to the low resolution encoder.

\boldparagraph{ResNet Architecture} For the ResNet-based encoder architecture~\cite{He2016CVPR}, we use the PyTorch ResNet-18 model pre-trained on ImageNet by removing the last linear layer. We also add an average pooling layer at the end to adjust the final channel size. Specifically, the pre-trained ResNet-18 architecture has $512$ channels in the last convolutional layer. We add the average pooling layer to reduce it to the desired output size
which is $3 \times 10$ and $4 \times 8$ for KITTI and Cityscapes, respectively. The decoder architecture is the symmetric version of the encoder which contains an extra transposed convolution at the end.

\boldparagraph{Latent Distributions} For each latent distribution, we use three ConvLSTMs for the prior, the posterior, and the predictor. The static prior and posterior take depth and pose features from the respective time-steps as input and output a latent distribution while the dynamic prior and posterior take flow features from the respective time-steps as input. The static predictor takes the previous time-step's depth and pose features and the latent variable from the posterior (or from the prior at the inference time) and predicts the static features of the future frames. The dynamic predictor takes previous time-step's flow features and the latent variable from the posterior (or from the prior at the inference time) and predicts the dynamic features of the future frames.
At the end, the predicted static features are decoded to depth and pose, and the dynamic features to optical flow. 
For the conditional model, we use the same static features to predict the residual flow.

\boldparagraph{Training Details}  We train our models for 150K iterations with a batch size of 16, or equivalent (by keeping the total number of seen video sequences the same, i.e. 150K $\times$ 16). We set the learning rate to $10 ^ {-4}$ and use the $\sigma$ loss, which is introduced in \cite{Rybkin2021ICML}, and learn the KLD weighting terms as it is proposed in \citet{Rybkin2021ICML}, for more details, please see the original paper. For KITTI, we use the VGG architecture and apply reconstruction losses to all of the branches, static, dynamic and final prediction. However, for Cityscapes we use the ResNet architecture and apply the reconstruction loss only to the final prediction to learn more disentangled static and dynamic reconstructions. We use a single Tesla V100 GPU for training. One training takes approximately 4-5 days.

\subsection{Baselines}
\boldparagraph{SVG~\cite{Denton2018ICML}}
For SVG, we utilize the VGG-based encoder and decoder architectures shown in \tabref{tab:enc} and \tabref{tab:dec}. We also change all the LSTMs to ConvLSTMs which use $3 \times 3$ convolutions operating on $3 \times 10$ or $4 \times 8$ feature maps for KITTI and Cityscapes, respectively.
We use $128 \times 3 \times 10$ feature maps for the output of the encoder architecture and $32 \times 3 \times 10$ channels for the latent variables, $z$. We train the architecture for 200K iterations with batch size of 10 and learning rate of $3\times 10^{-4}$. We use a single Tesla T4 GPU for the training. The training takes approximately 1-2 days.

\boldparagraph{SRVP~\cite{Franceschi2020ICML}}
For SRVP, we employ the same architectures used in SVG, which are shown in \tabref{tab:enc} and \tabref{tab:dec}. However, we add a pooling layer at the end to convert the output feature map a vector because the SRVP architecture uses MLPs to process the output of the encoder and the decoder. We use $128 \times 1 \times 1$ feature map, which is pooled from the feature map of $128 \times 3 \times 10$,  for the output of the encoder, and $32 \times 1 \times 1$ for the latent variables. We use $2$ Euler steps, skip connections from the encoder to the decoder, $3$ frames for the content variable, $0.2$ for the variance of negative log-likelihood loss for the reconstruction. We train the model for 100K iterations with the batch size of 24 and learning rate of $3\times 10^{-4}$ . We use a single Tesla V100 GPU for training. The training takes approximately 1-2 days.

\boldparagraph{Improved-VRNN~\cite{Castrejon2019ICCV}}
For the Improved-VRNN, we resized the images to $304 \times 96$ on KITTI which is divisible by 16 as required by their architecture to obtain the closest resolution to ours, and trained for $390\text{K}$ iterations with a batch size of $3$. On Cityscapes, we resized the images to the same resolution as ours, \ie $256 \times 128$ and trained for $360\text{K}$ iterations, using a batch size of $3$. The training takes approximately 5-6 days on a single V100 GPU. Because of large memory usage of the model, we could not use a larger batch size or train any longer.

\section{Additional Results}
\label{sec:results}

\subsubsection{Quantitative Results Across Time}
\begin{sidewaysfigure}[t]
    \begin{subfigure}{\linewidth}
        \centering
        \includegraphics[width=.33\textwidth]{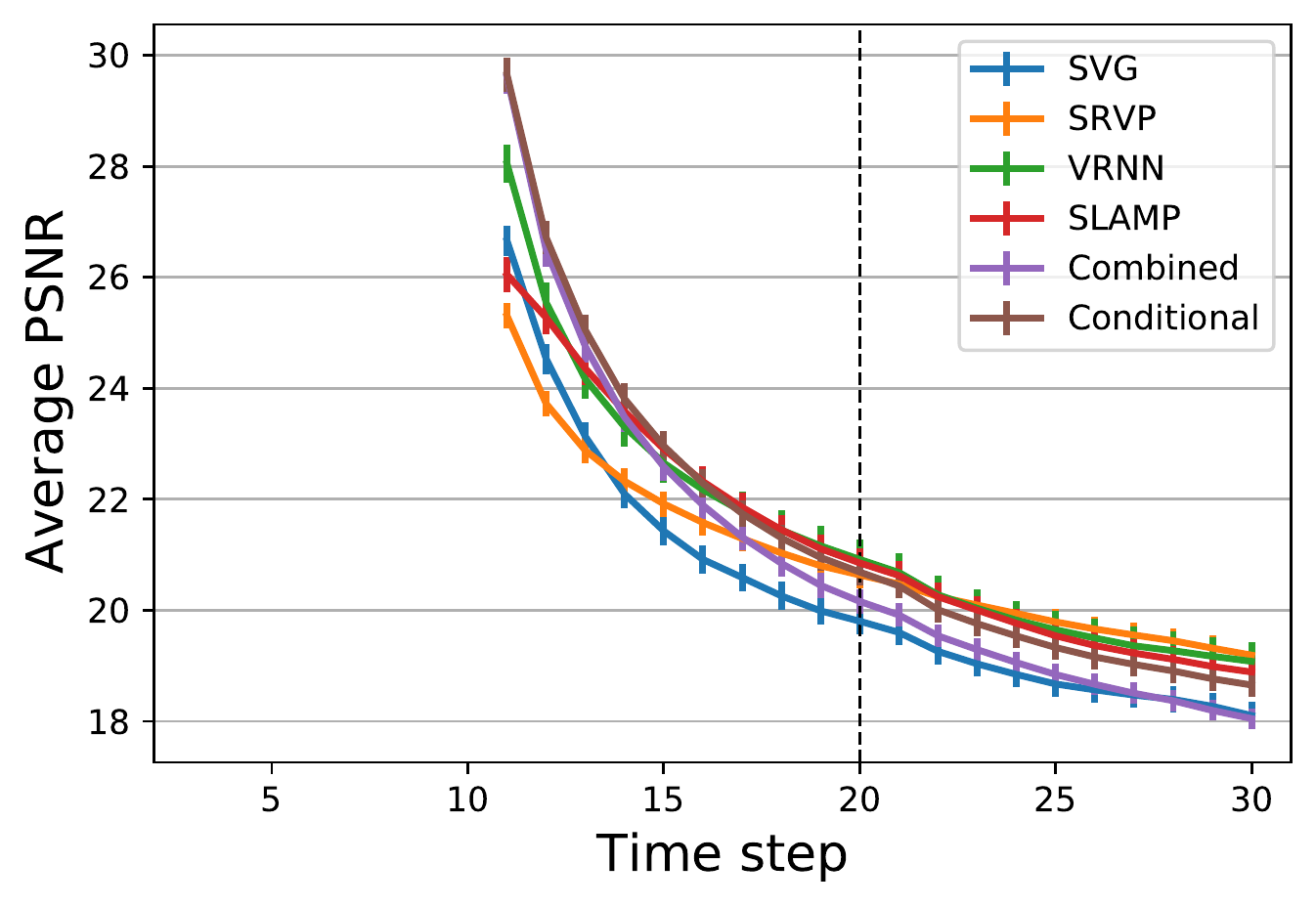}\hfill
        \includegraphics[width=.33\textwidth]{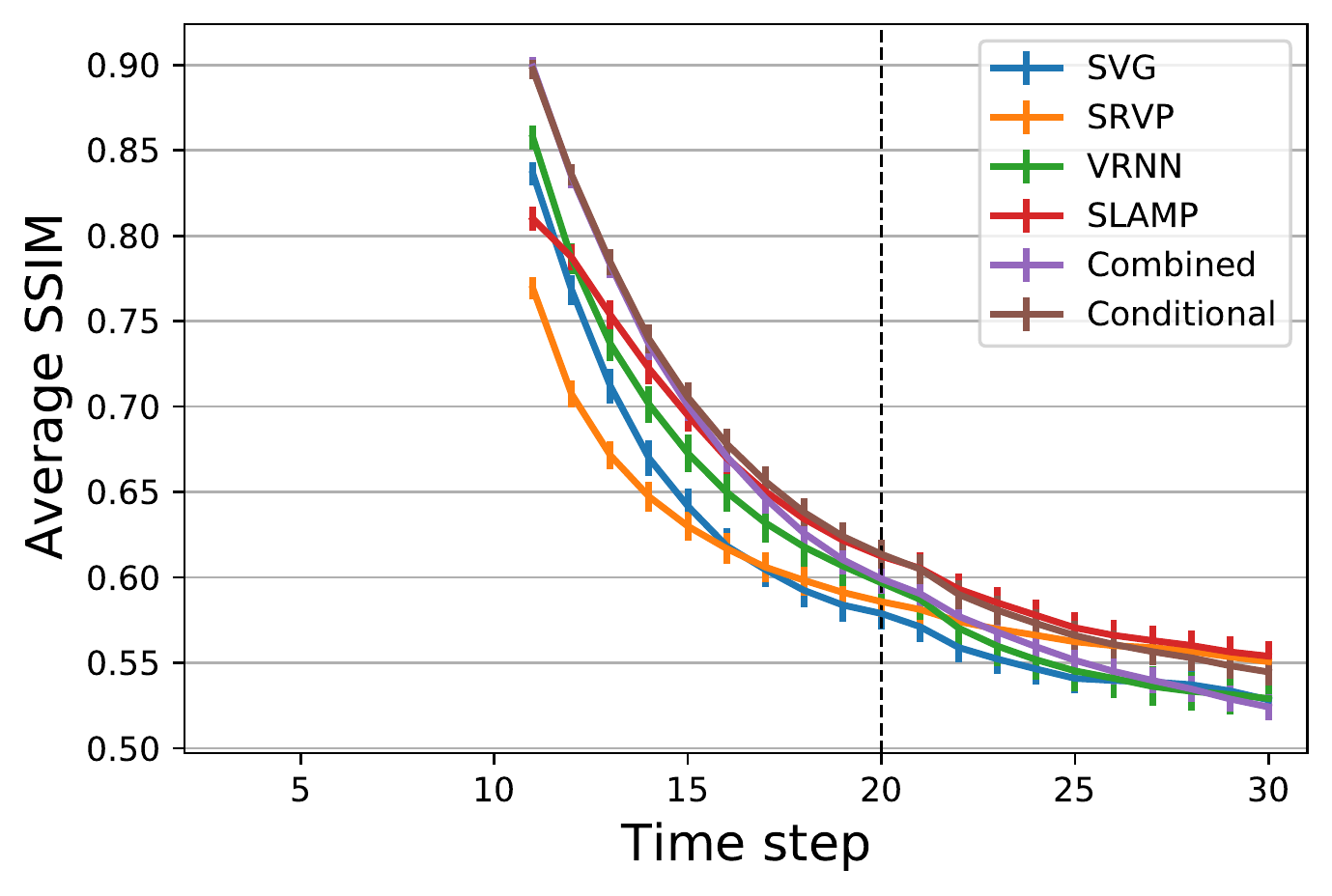}\hfill
        \includegraphics[width=.33\textwidth]{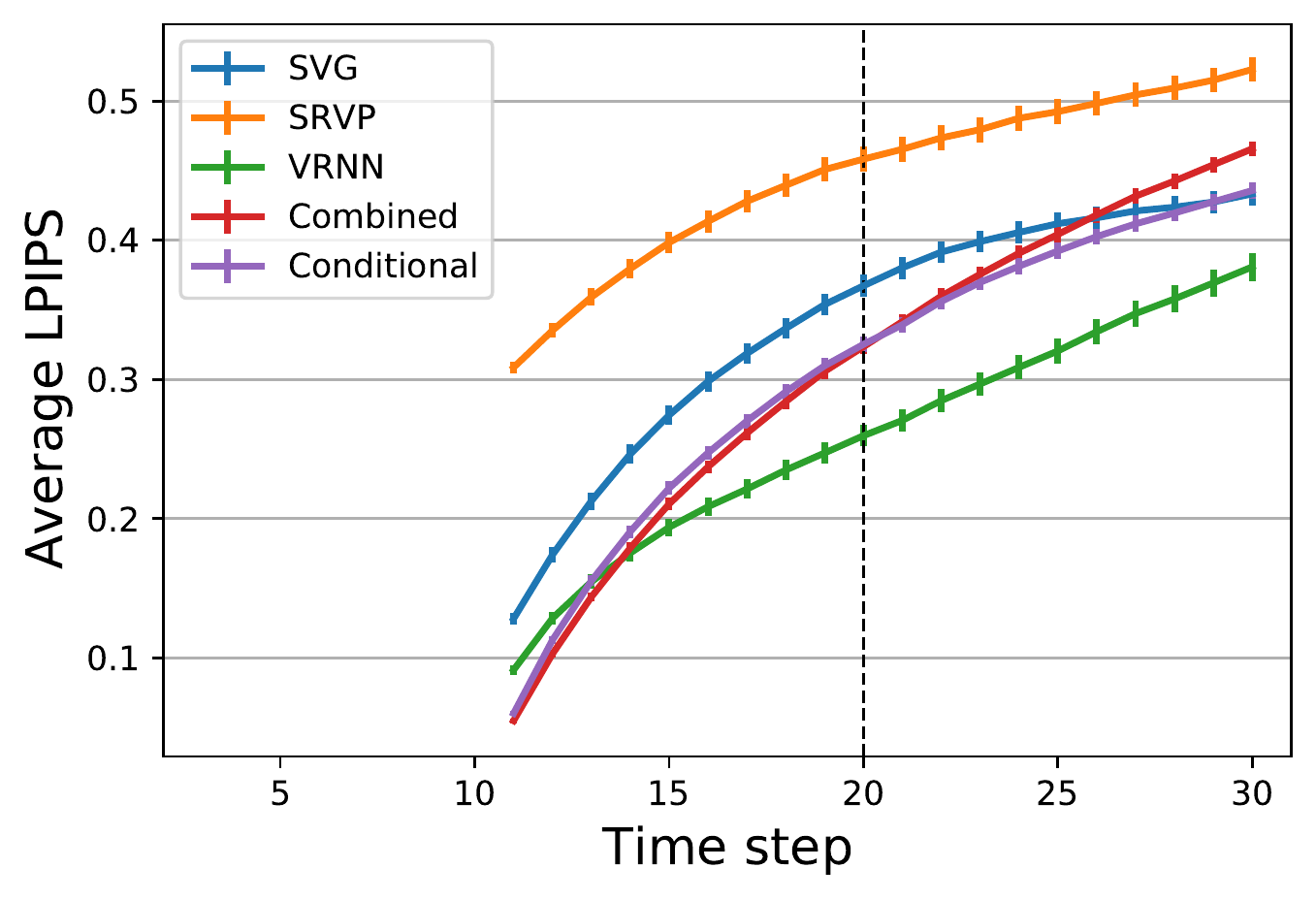}
        \caption{Cityscapes}
        \label{fig:city_time}
    \end{subfigure}
\\
    \begin{subfigure}{\linewidth}
        \centering
        \includegraphics[width=.33\textwidth]{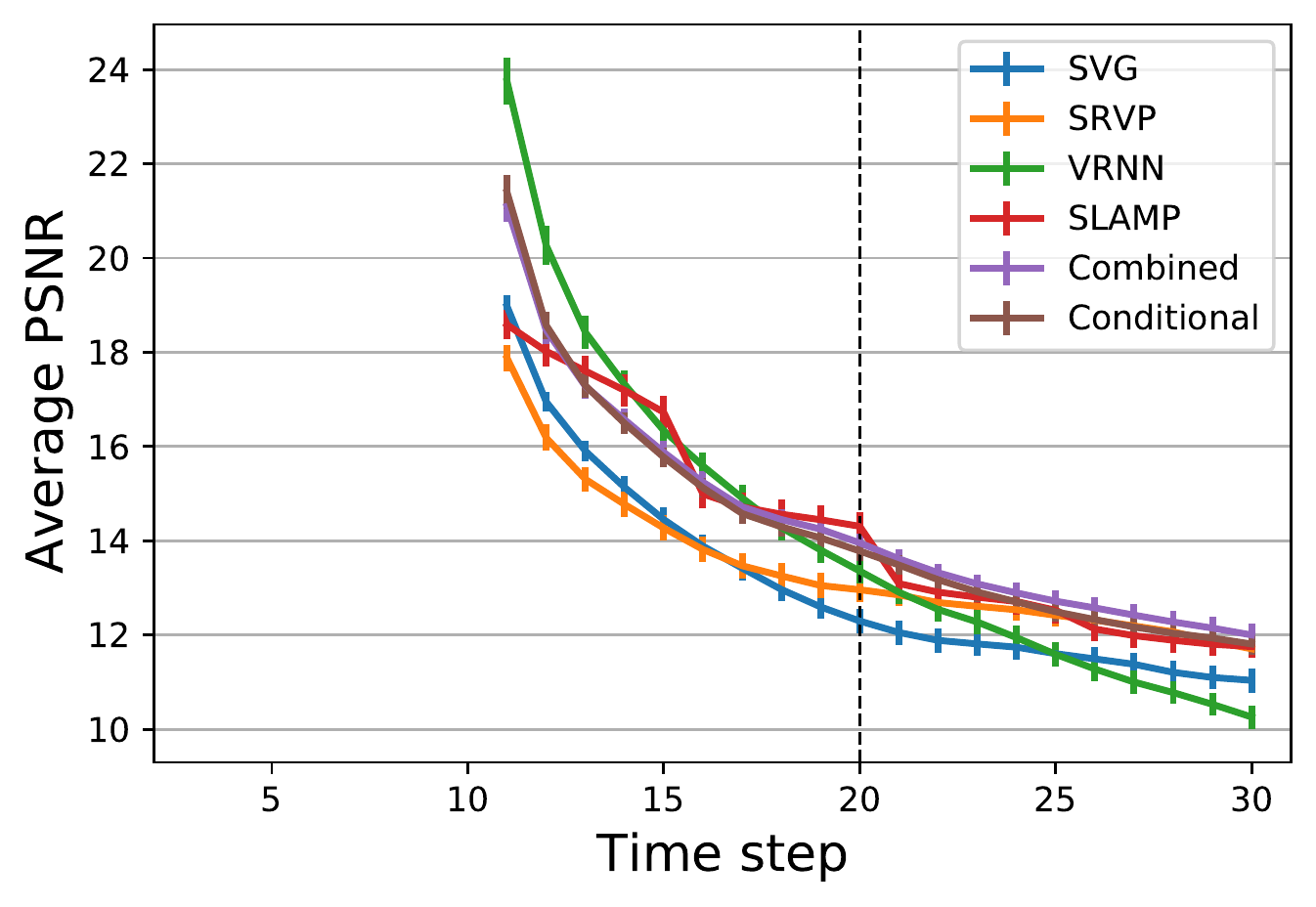}\hfill
        \includegraphics[width=.33\textwidth]{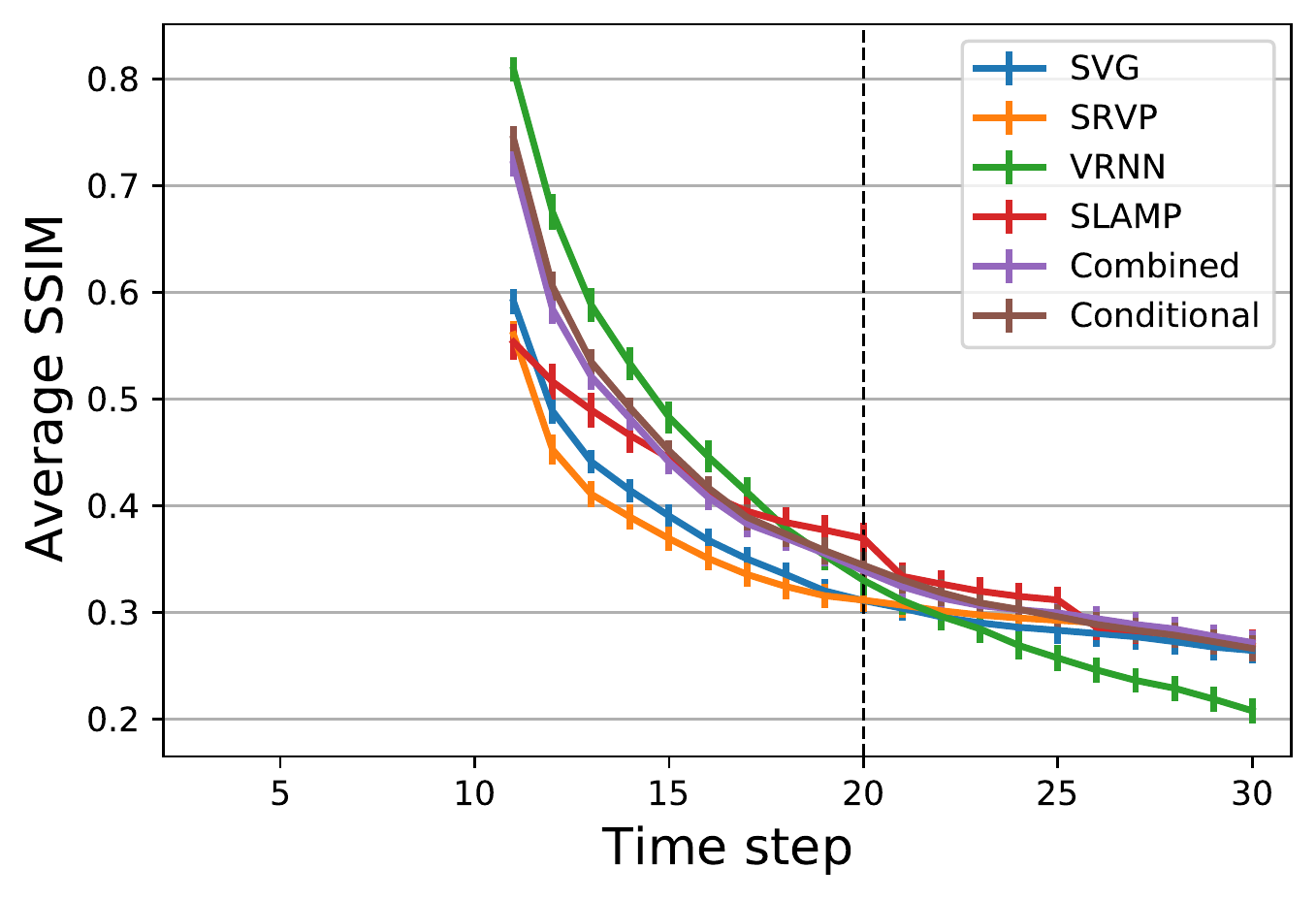}\hfill
        \includegraphics[width=.33\textwidth]{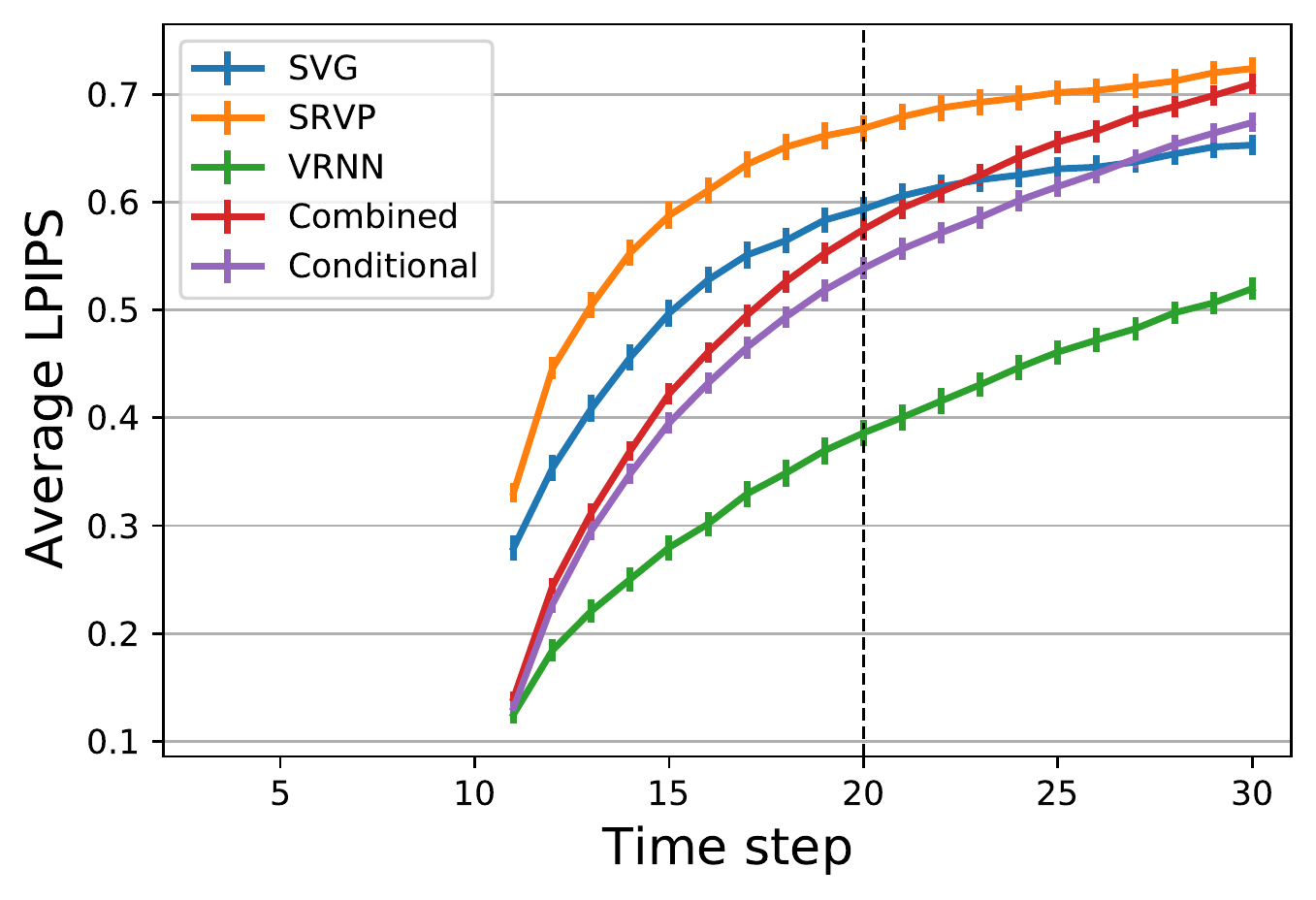}
        \caption{KITTI}
        \label{fig:kitti_time}
    \end{subfigure}
\caption{\textbf{Performance Across Time.} We plot each metric across time-steps for different methods on Cityscapes (\subref{fig:city_time}) and KITTI (\subref{fig:kitti_time}).}
\label{fig:quan_time}
\end{sidewaysfigure}

We provide the plots of the results in the main text across time-steps in \figref{fig:quan_time}. While the performances of competing models degrade in time, our models, especially the \textit{Conditional}, achieve better overall performances, along with~Improved-VRNN~\cite{Castrejon2019ICCV}.

\subsection{Additional Qualitative Results}
\begin{sidewaysfigure}[t]
\centering
\includegraphics[width=\textwidth]{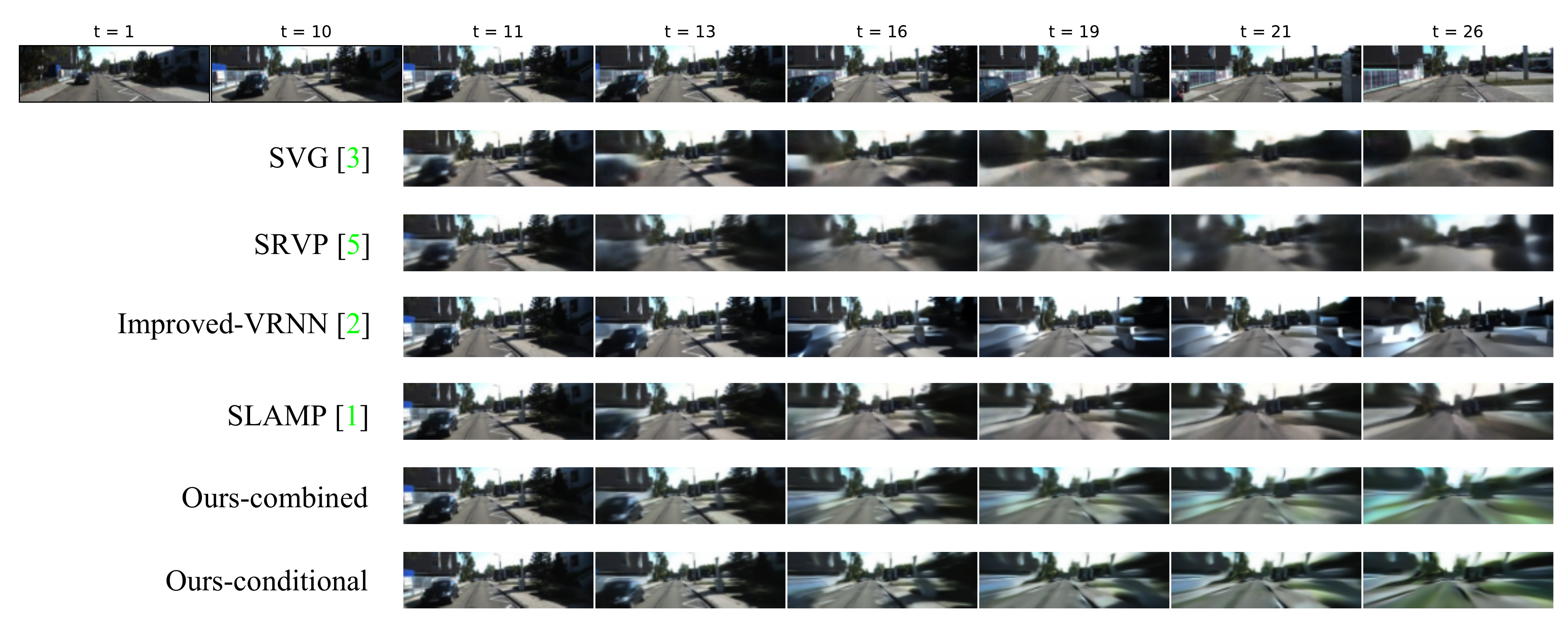}
\caption{\textbf{Comparison to Other Methods on KITTI.} Our models perform similarly to Improved-VRNN and outperform SVG and SRVP, especially for further time-steps that are harder to predict.}
\label{fig:kitti_comp1}
\end{sidewaysfigure}

\begin{sidewaysfigure}[t]
\centering
\includegraphics[width=\textwidth]{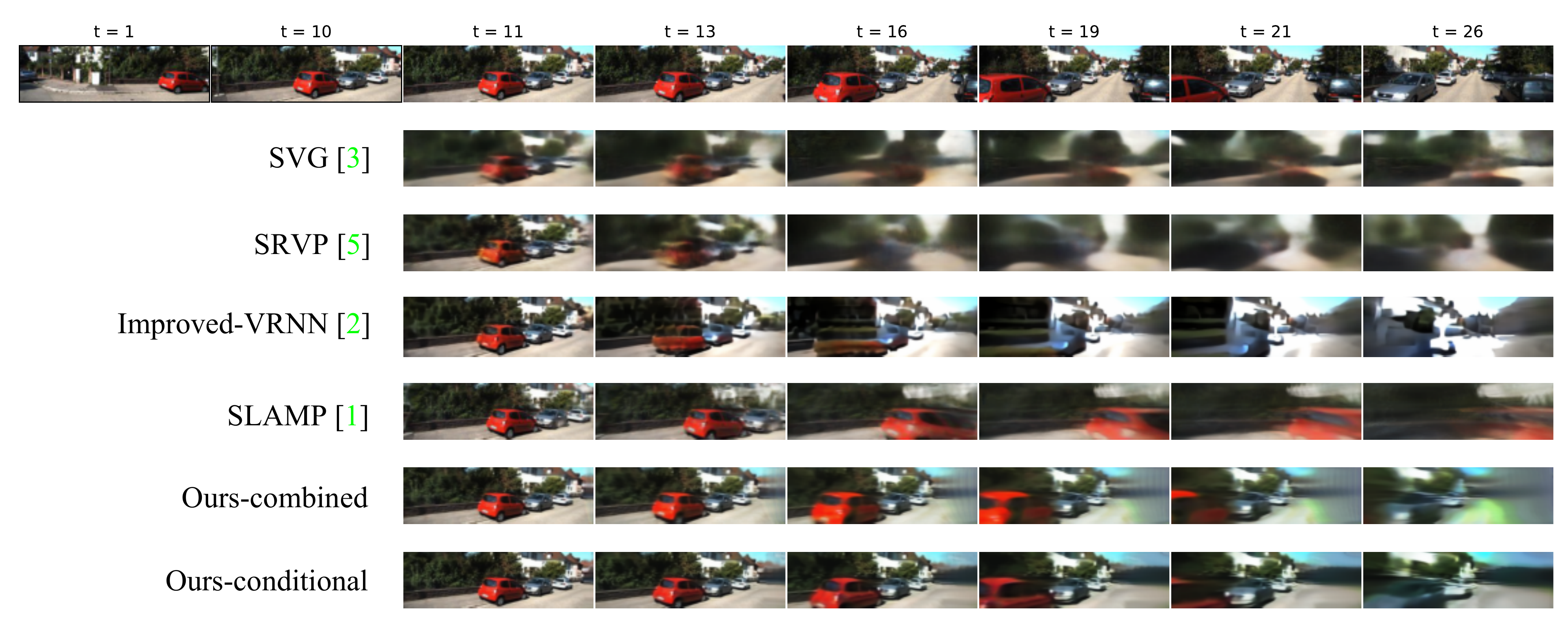}
\caption{\textbf{Comparison to Other Methods on KITTI.} This example shows a challenging turning-right case where only our models can predict the appearance of the red car in later time-steps.}
\label{fig:kitti_comp2}
\end{sidewaysfigure}

\begin{sidewaysfigure}[t]
\centering
\includegraphics[width=\textwidth]{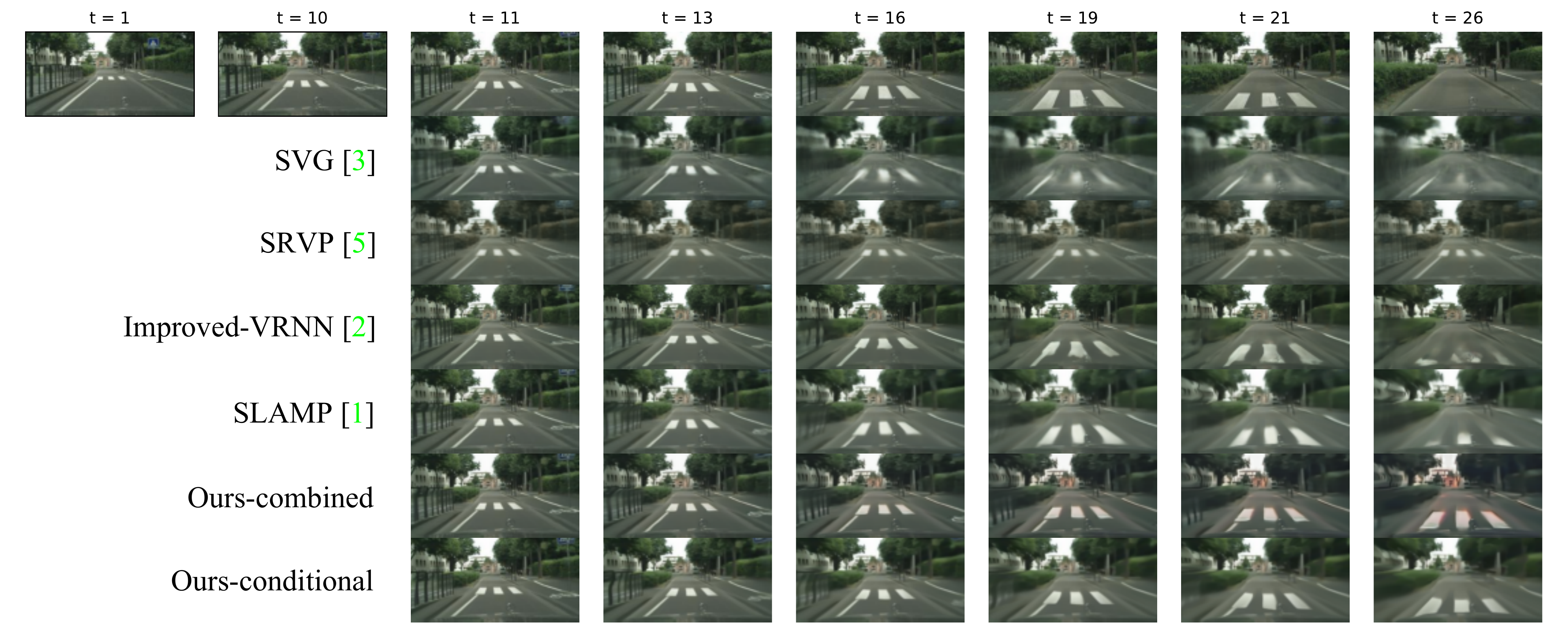}
\caption{\textbf{Comparison to Other Methods on Cityscapes.} This example shows a case with only ego-motion where our models most accurately predict the effect of perspective projection on the lane markings.}
\label{fig:city_comp1}
\end{sidewaysfigure}

\begin{sidewaysfigure}[t]
\centering
\includegraphics[width=\textwidth]{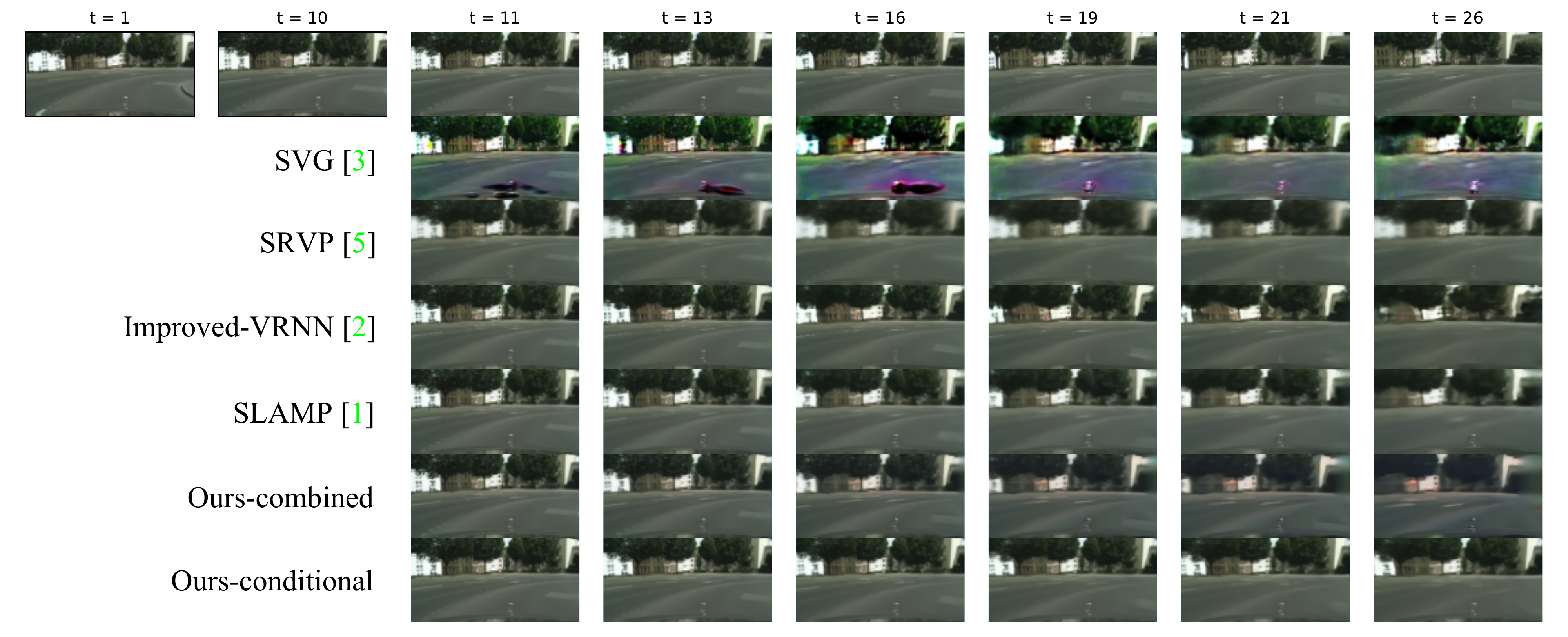}
\caption{\textbf{Comparison to Other Methods on Cityscapes.} This examples shows another challenging turning case which can only be handled by Improved-VRNN and our models.}
\label{fig:city_comp2}
\end{sidewaysfigure}

\begin{sidewaysfigure}[t]
\centering
\includegraphics[width=\textwidth]{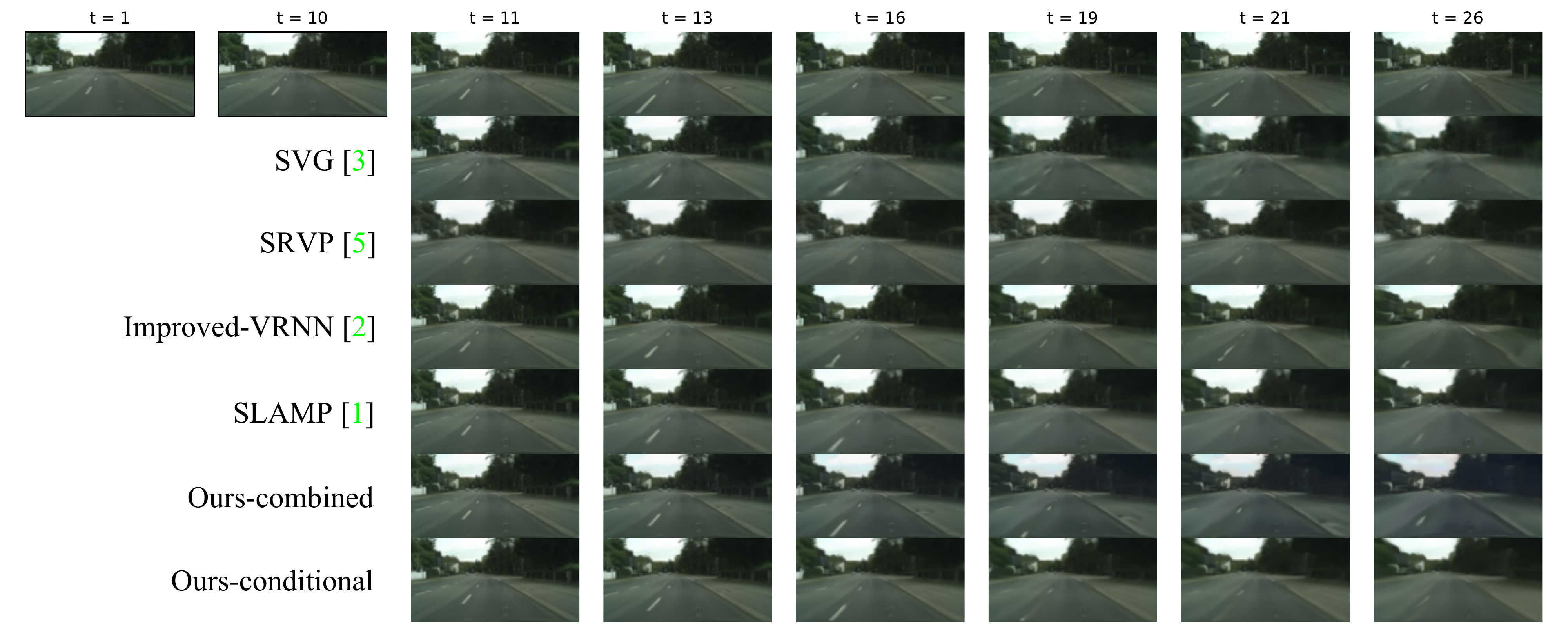}
\caption{\textbf{Comparison to Other Methods on Cityscapes.} This figure shows another example of modelling the effect of perspective projection in the scene. Our models can better preserve the side walk compared to Improved-VRNN which imagines another lane on the sidewalk.}
\label{fig:city_comp3}
\end{sidewaysfigure}

\begin{sidewaysfigure}[t]
\centering
\includegraphics[width=\textwidth]{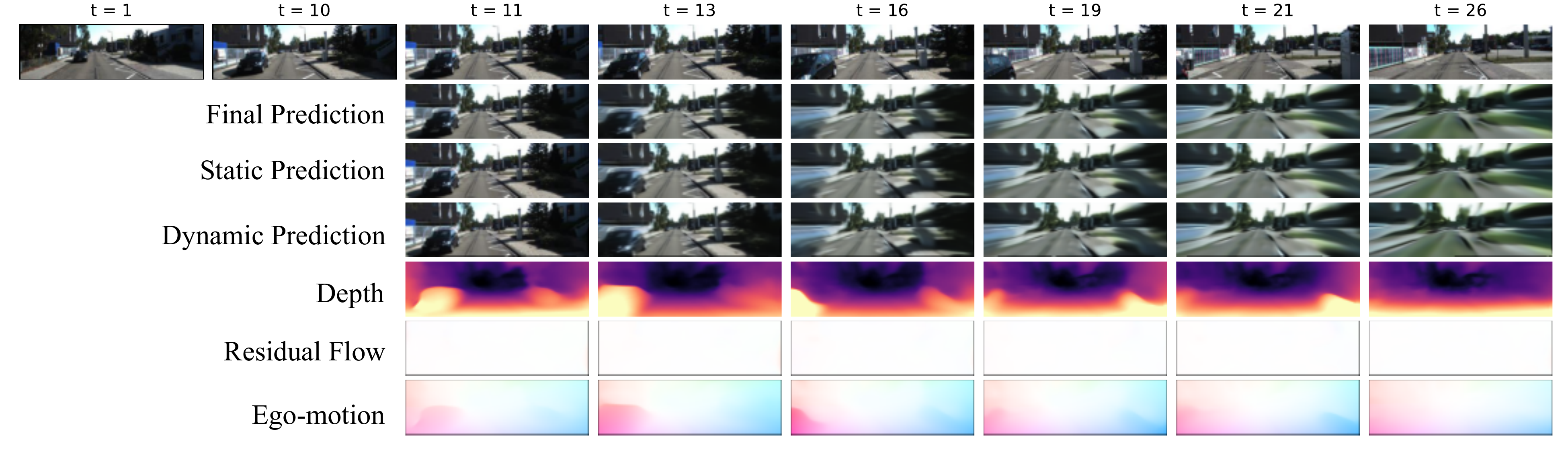}
\caption{\textbf{Detailed Visualization of our Conditional Model on KITTI.} We show the final reconstruction as well as the reconstructions from static and dynamic parts separately. As can be seen from the following visualizations of depth, residual flow, and flow due to ego-motion, most of the motion in the scene can be captured by the ego-motion.}
\label{fig:kitti_detailed1}
\end{sidewaysfigure}

\begin{sidewaysfigure}[t]
\centering
\includegraphics[width=\textwidth]{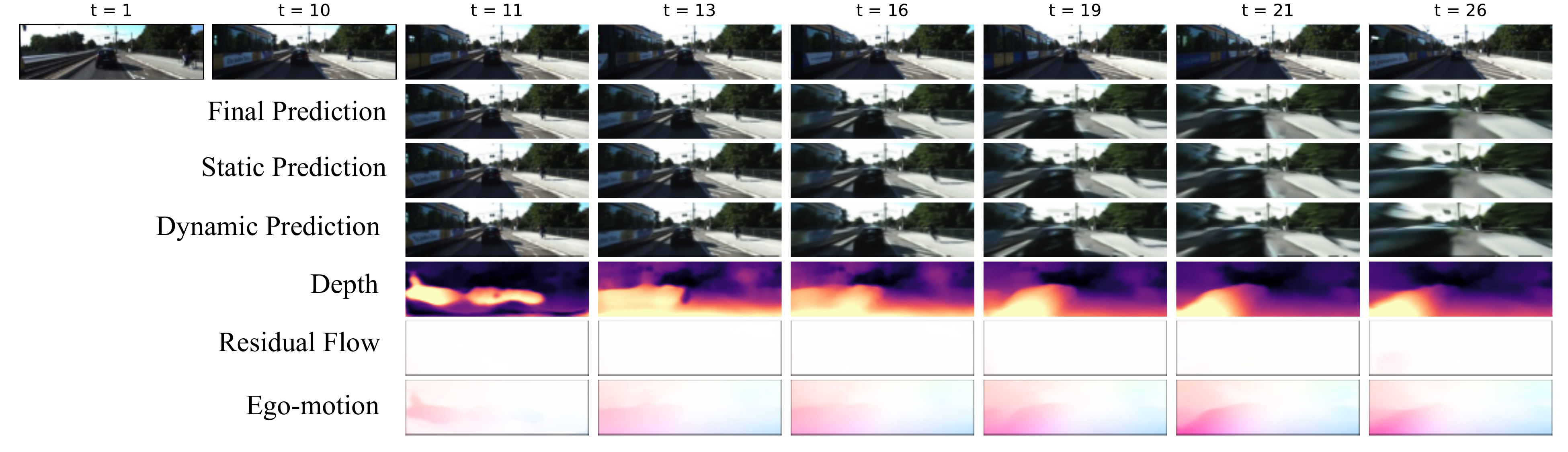}
\caption{\textbf{Detailed Visualization of our Conditional Model on KITTI.} We show the final reconstruction as well as the reconstructions from static and dynamic parts separately. As can be seen from the following visualizations of depth, residual flow, and flow due to ego-motion, most of the motion in the scene can be captured by the ego-motion.}
\label{fig:kitti_detailed2}
\end{sidewaysfigure}

\begin{sidewaysfigure}[t]
\centering
\includegraphics[width=\textwidth]{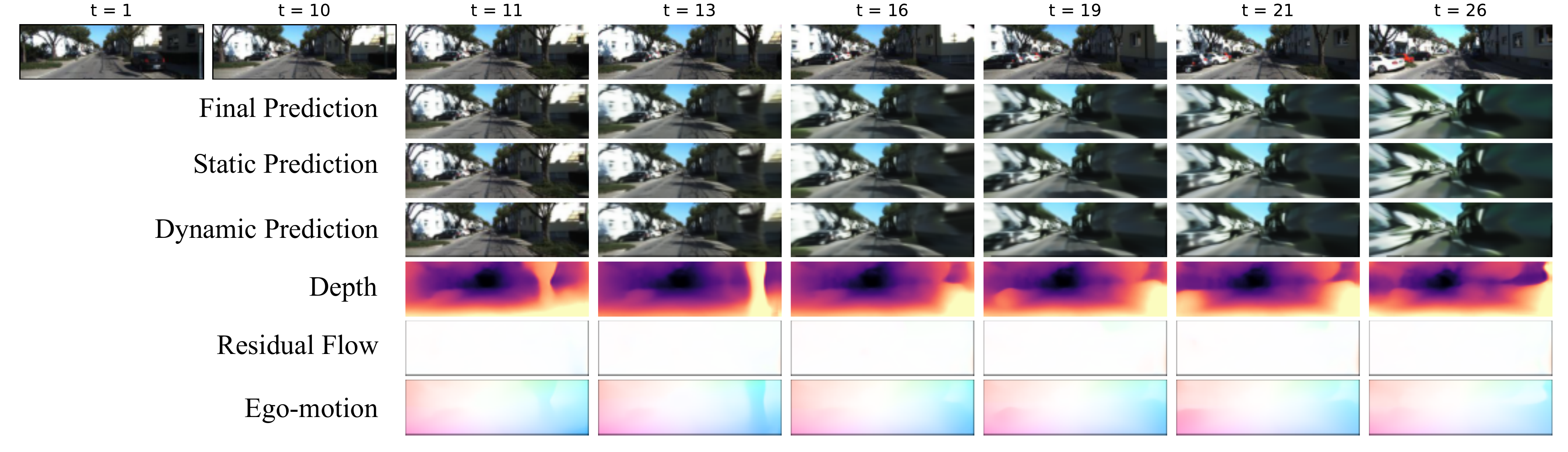}
\caption{\textbf{Detailed Visualization of our Conditional Model on KITTI.} We show the final reconstruction as well as the reconstructions from static and dynamic parts separately. As can be seen from the following visualizations of depth, residual flow, and flow due to ego-motion, most of the motion in the scene can be captured by the ego-motion.}
\label{fig:kitti_detailed3}
\end{sidewaysfigure}

\begin{sidewaysfigure}[!t]
\centering
\begin{subfigure}[b]{\textwidth}
   \includegraphics[width=1\linewidth]{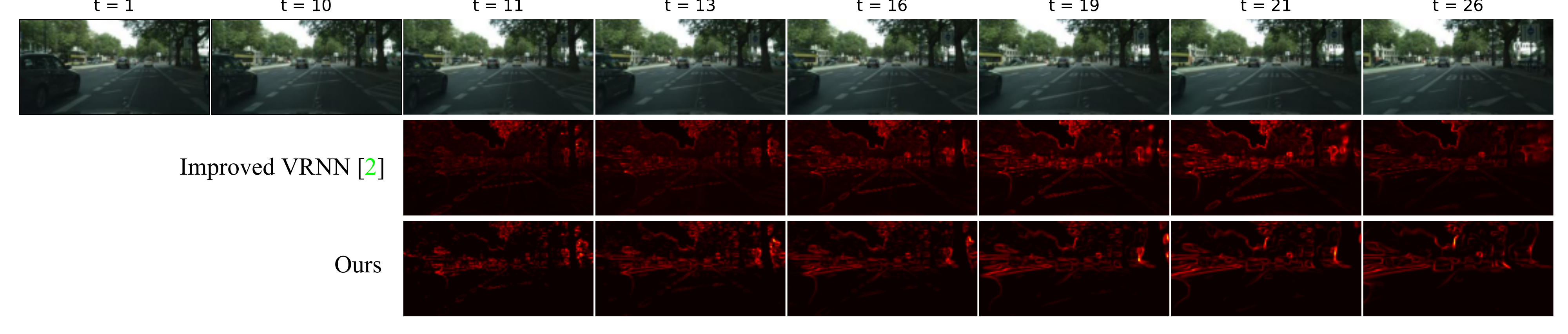}
\end{subfigure}
\begin{subfigure}[b]{\textwidth}
   \includegraphics[width=1\linewidth]{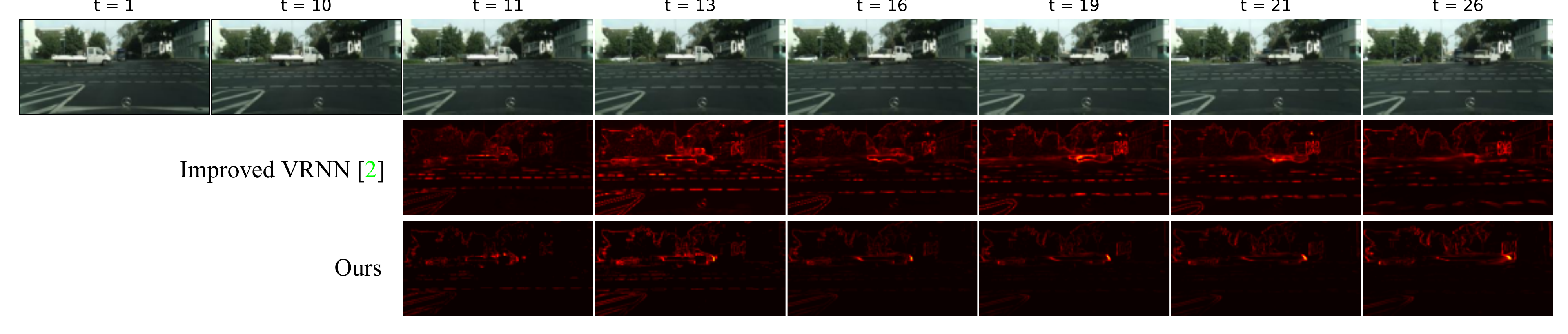}
\end{subfigure}
\begin{subfigure}[b]{\textwidth}
   \includegraphics[width=1\linewidth]{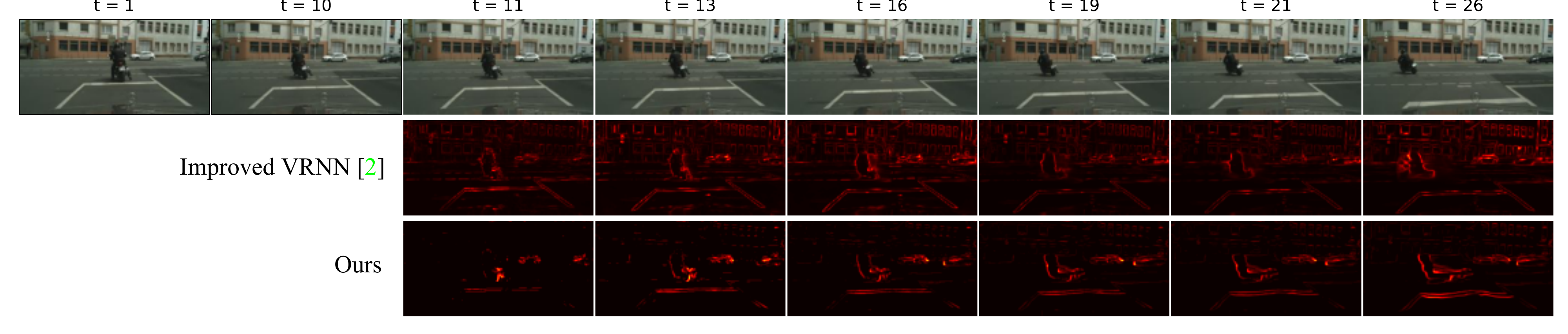}
\end{subfigure}
\caption{\textbf{Visualization of Diversity.} This figure shows more examples of diversity visualized as the standard deviation of samples generated. As can bee seen from three examples, our model consistently can pinpoint uncertainty to foreground object regions whereas it is more equally distributed across the image for Improved-VRNN.}
\label{fig:diversity_sup}
\end{sidewaysfigure}

In this section, we provide additional qualitative comparisons to other methods on both KITTI and Cityscapes as well as more examples of detailed predictions on KITTI and diversity visualizations on Cityscapes. Note that we attach the gif versions of these results for an easier comparison. 
We compare both our models \textit{Combined} and \textit{Conditional} to other methods including SVG~\cite{Denton2018ICML}, SRVP~\cite{Franceschi2020ICML}, and Improved-VRNN~\cite{Castrejon2019ICCV} on KITTI in \figref{fig:kitti_comp1} and \figref{fig:kitti_comp2}, and on Cityscapes in \figref{fig:city_comp1}, \figref{fig:city_comp2}, and \figref{fig:city_comp3}.
We also provide additional plots with detailed visualizations of our results including the depth prediction, the residual flow, and flow due to ego-motion in \figref{fig:kitti_detailed1}, \figref{fig:kitti_detailed2} and \figref{fig:kitti_detailed3}. 
Lastly, we provide additional visualizations of the standard deviation maps as an indication of diversity of predictions in \figref{fig:diversity_sup}. As stated in the main paper, our model focuses on the moving objects as a source of stochasticity rather than the whole scene as in the case of Improved-VRNN.

\end{appendices}




\vfill


\end{document}